\documentclass[sigconf]{acmart}

\usepackage{booktabs} 
\usepackage{balance}
\usepackage{filecontents}
\usepackage{xurl}
\usepackage{xcolor}
\usepackage{subfig}
\usepackage{tabularx}
\newcolumntype{Y}{>{\centering\arraybackslash}X}

\usepackage{enumerate}
\usepackage{enumitem}
\usepackage[normalem]{ulem}
\usepackage{float}
\usepackage{multicol}
\usepackage{multirow}

\usepackage{tabularx}

\begin{document}
\balance
\title[Out-of-Context Language Modelling]{``TL;DR:'' Out-of-Context Adversarial Text Summarization and Hashtag Recommendation}


\author{Peter Jachim}
\affiliation{%
  \institution{DePaul University}
  \streetaddress{243 S Wabash Ave}
  \city{Chicago}
  \state{IL}
  \postcode{60604}
}
\email{pjachim@depaul.edu}

\author{Filipo Sharevski}
\affiliation{%
  \institution{DePaul University}
  \streetaddress{243 S Wabash Ave}
  \city{Chicago}
  \state{IL}
  \postcode{60604}
}
\email{fsharevs@cdm.depaul.edu}

\author{Emma Pieroni}
\affiliation{%
  \institution{DePaul University}
  \streetaddress{243 S Wabash Ave}
  \city{Chicago}
  \state{IL}
  \postcode{60604}
}
\email{epieroni@depaul.edu}

\renewcommand{\shortauthors}{P. Jachim, F. Sharevski, E. Pieroni}

\begin{abstract}
This paper presents \textit{Out-of-Context Summarizer}, a tool that takes arbitrary public news articles out of context by summarizing them to coherently fit either a liberal- or conservative-leaning agenda. The \textit{Out-of-Context Summarizer} also suggests hashtag keywords to bolster the polarization of the summary, in case one is inclined to take it to Twitter, Parler or other platforms for trolling. \textit{Out-of-Context Summarizer} achieved 79\% precision and 99\% recall when summarizing COVID-19 articles, 93\% precision and 93\% recall when summarizing politically-centered articles, and 87\% precision and 88\% recall when taking liberally-biased articles out of context. Summarizing valid sources instead of synthesizing fake text, the \textit{Out-of-Context Summarizer} could fairly pass the ``adversarial disclosure'' test, but we didn't take this easy route in our paper. Instead, we used the \textit{Out-of-Context Summarizer} to push the debate of potential misuse of automated text generation beyond the boilerplate text of responsible disclosure of adversarial language models.    

\end{abstract}



\settopmatter{printacmref=false}
\maketitle
\fancyhead{} 
\pagenumbering{gobble}


\section{Introduction}
The goal of online trolling is often assimilation into the public conscience by maintaining fake narratives with improbable interpretations while remaining undetected \cite{Sirivianos}. Yet this paper presents an alternative means of trolling, or trolling a content feeder, by extracting legitimate parts of public discourse from news articles, selectively isolating polarizing parts, and taking them \textit{out of context} in an effort to advance an adversarial agenda. Text summarization by \textit{Out-of-Context Summarizer}, the adversarial language model presented in this paper, is the natural means to this end, as readers are often short on time and prefer the soundbite of issues in ``tl;dr'' format. The simplicity associated with text summarization is preferable to lengthy readings, and therefore leaves the technique ripe for adversarial misuse. 

Social media has become an outlet for news, and users on sites like Twitter or Parler rely on short pieces of text in order to understand polarizing topics or narratives. Text summaries of news articles are continuous feeds of fresh content that can be used by adversaries in a variety of different ways, including as newsletters, or continuous Twitter or Parler content. If an adversary were to upload a summary to Twitter, it would need to fall within the character limit  or use a conversational thread with few consecutive tweets \cite{Twitter}. Similarly, Parler deploys a ``Read More..." functionality which deters the use of messages longer than 1000 characters \cite{chi2021}. Studies suggest that when users are scrolling social media, particularly looking for political content, they are conditioned to respond well to relatively short messages or threats that evoke a feeling of being informed \cite{Boukes, Geeng}. Even more, people tend to take content from trusted sources like legitimate news articles at face value as a heuristic that saves time \cite{Zappavigna}. Text summarization is able to satiate the need for speedy news consumption, by giving users the ``tl;dr" critical information from an article, so that a user can keep scrolling. 

Summaries are usually accompanied with embedded features like links or hashtags in order to facilitate content diffusion to reach a broad audience in a crowded social media landscape \cite{Pancer}. Understanding that this increases audiences' exposure, adversaries were quick to diffuse alternative narratives on social media including polarizing hashtags and links to posts on less regulated platforms like 4chan \cite{Twitter, nspw2020}. The addition of links is to avoid hard moderation by mainstream platforms for openly disseminating trolling content. The particular addition of hashtags, which help to organize content and track discussions based on keywords, is not to avoid moderation but to yield inferences about the summaries that reinforce the polarizing positions \cite{Zappavigna}. In anticipation of an adversary utilizing the approach to bolster a position, we have also designed \textit{Out-of-Context Summarizer} to predict potential hashtag keywords to accompany the suggested out-of-context summary. 

Often, in trying to understand trolling online or detect it, researchers utilize complex automated language models \cite{Capistrano, Fornacciari}. While these models certainly have the potential for adversarial misuse, the complexity associated provides little incentive for trolls. Yet the text summarization method used in \textit{Out-Of Context Summarizer} provides a relatively quick and efficient way to generate adversarial narratives from already existing pieces of public discourse like news articles. Using machine learning, \textit{Out-of-Context Summarizer} ``borrows'' the experience of journalists who understand their audiences in order to weaponize that relationship, by swaying the summarization to seem either left or right-leaning, without raising alarm from users who trust the new sources they read. After generating alternative narratives under the guise of legitimate journalism, trolls can upload summaries quickly to social media sites like Twitter or Parler, to change the context of the narrative. By manipulating the summarization to favor either conservative or liberal sentences, the perspective of the polarizing topic is manipulated, but remains undetectable, as the selected sentences are actually present in publicly available news articles. Traditional trolling may be more easily detectable, as the alternative narratives are often known and also widely dispelled, but by advancing adversarial agendas through legitimate pieces of public discourse, trolling becomes much harder to identify and poses a security risk worth disclosing. 


\section{Automatic Text Summarization} \label{sec:ats}
Automatically condensing large texts to contain few important sentences has been, and remains, an imperative of many language models in a sincere effort to maximize the entropy of information delivered to a user for consumption in minimum required time \cite{Mahak, Jones}. The sheer volume of online text content in particular makes it hard for users to sift through textual data that is often redundant, unverified, and incorrectly formatted. Robust automatic text summarization systems, therefore, received a widespread attention from the natural language processing and artificial intelligence communities. These systems usually are classified based on the language, input size, summarization approach, nature of the output summary, summarization algorithm, summary content, summarization domain, language, and type of summary \cite{El-Kassas}. 

The \textit{Out-of-Context Summarizer} manipulates a single neutral input article, based on a machine learning model trained on a corpus of polarized texts, to create a polarized article summary, as described in more detail in Section \ref{sec:ats}. The modular design of the system allows the adversary to quickly change the contexts of the system to target different articles in different contexts, or even different languages. The summarization approach is \textit{extractive}---our \textit{Out-of-Context Summarizer} selects sentences from the input article and then concatenates them to form the summary. In Section \ref{sec:discussion} we expand on the future adaptation for generating \textit{abstractive} summaries, by using sentences that are different than the original sentences. This adaptation is possible given that approaches for automated word manipulation in various forms of online text such as email \cite{eurosp2020}, Twitter posts \cite{nspw2020}, Wikipedia articles \cite{acsac2020}, or text converted to speech by an voice assistant \cite{malexa2021} exist. 

The nature of the the \textit{Out-of-Context Summarizer} output is generic, but the hashtag generator can be repurposed to be used for querying particular keywords for appending the generic output, as shown in Section \ref{sec:examples}. The summary output of \textit{Out-of-Context Summarizer} algorithm is both \textit{indicative}, alerting the user about the source content, and specifically \textit{informative}, outlining the main contents of the original text (though with an original twist, which makes the \textit{Out-of-Context Summarizer} appealing for adversarial text summarization). \textit{Out-of-Context Summarizer} can produce various lengths of summaries, including headlines, sentence-level, highlights, or a full summary \cite{Dernoncourt}. Although we focus on full-text summaries, in Section \ref{sec:discussion} we provide examples of more abbreviated outputs. We also focus on a specific domain of temporary events like elections or the COVID-19 pandemic, but \textit{Out-of-Context Summarizer} could be customized to produce summaries from different domains, for example incorporating elements from medical documents or journal papers to supplement the COVID-19 summaries. We also focus on creating alternative narratives out of context from news articles, but one might use \textit{Out-of-Context Summarizer} for example to summarize a legislative bill for the same purposes. 

\section{Adversarial Text Summarization} \label{sec:ats}
The scope of our adversarial text summarization mechanism includes: the gathering and structuring of data, data modelling, and text summarization aided by analysis of the fitted model. Figure 1 shows the high-level overview of the language model behind the \textit{Out-of-Context Summarizer}. The out-of-context summarization starts when a website of interest publishes an article. The text from the article is then scraped and formatted into a cohesive, labelled dataset. The labelled dataset is passed to a script that fits a machine learning pipeline to the labelled dataset, and saves the fitted pipeline. Finally, the text summarization process takes the fitted model, analyzes the prediction probability to decide the out-of-context weights, and creates text summaries that can be passed along to the delivery mechanism. The sources used can be adjusted, and new scrapers can be created to enable the adversary to choose from a wide range of online news outlets. 

\begin{figure*}[htpb]
\includegraphics[width=0.58\textwidth]{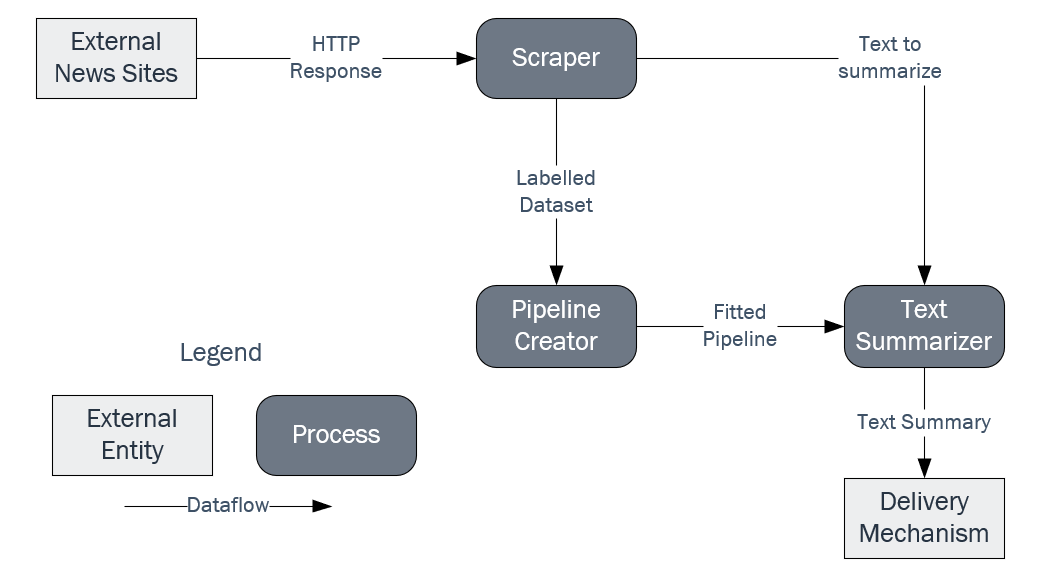}
\caption{\textit{Out-of-Context Summarizer} Dataflow diagram showing how data flows through the Out-of-Context Summarizer, from the original text in the external news site to the delivery mechanism.}
\end{figure*}

For the purposes of this paper, we trained the classification model used in the \textit{Out-of-Context Summarizer} on articles from \textit{Vox} and \textit{Breitbart} initially scraped in February of 2021. We selected these two websites to establish liberal and conservative baselines, respectively \cite{Wihbey}, but future adaptations could select text from any number of news sources. To demonstrate the \textit{Out-of-Context Summarizer} capabilities, we generated three adversarial text summaries in three different contexts: first, we summarized the articles included in \textit{The Hill} as a politically-centered news outlet based on the learned data from \textit{Vox} and \textit{Breitbart}; next, we repeated the experiment on March $10^{th}$ of 2021 but with a limited scope of the articles of interest to the COVID-19 pandemic; and lastly, we tried creating text summaries from \textit{BuzzFeed}, while \textit{Out-Of-Context Summarizer} was again trained on text scraped from \textit{Vox} and \textit{Breitbart}.

The choice of the articles was to provide for a balanced editorial input with \textit{Breitbart} on the political right in the U.S., \textit{Vox} and \textit{Buzzfeed} on the political left, and \textit{The Hill} at the center \cite{Freelon}. This was a deliberate choice but for adversarial purposes, the articles can be selected from different sources – not just on the political spectrum but on any polarizing editorial stance. We deliberately avoided using mainstream outlets like Fox News, New York Times, or MSNBC to demonstrate the versatility of the out-of-context language modeling, but also to showcase summarizes that could not be immediately associated with the tune of a mainstream outlet. For the language modeling, we must note, an adversary might not be using news articles per se; they can train the \textit{Out-of-Context Summarizer} on the textual portion of available Twitter information operations datasets \cite{Twitter}, for example on a particular topic like COVID-19 \cite{nspw2020}.

\subsection{Web Scraping}
To obtain the data, we scraped four sites on the political spectrum: \textit{The Hill}, \textit{Buzzfeed}, \textit{Vox}, and \textit{Breitbart} (shown in Table \ref{tab:scrap}). To scrape the data, in our baseline we used a python script that leveraged the requests and \texttt{bs4 (Beautiful Soup 4)} libraries, to extract and parse the HTML for each site. To find the links, we found high level pages that linked to each of the articles we were interested in, collected the links, and used those links to figure out which text we needed to collect. Due to the context in which we built the \textit{Out-of-Context Summarizer}, we decided that it was easier to manually delete long passages of JavaScript, and it would allow us to avoid repeatedly sending traffic to the \textit{Breitbart} and \textit{Vox} pages to scrape the data. An adversary could use regular expressions to avoid downloading the JavaScript in the first place with additional experimentation with the scraper. 

\begin{table}[!h]
    \centering
    \caption{\textit{Quantity of Scraped Data Across Sources}: \textit{Breitbart} and \textit{Vox} include the number of sentences because that is the unit of text used as a sample for the classification model.}
    \label{tab:scrap}
 \begin{tabularx}{\linewidth}{|Y|Y|Y|Y|Y|}
 \hline 
    \textbf{\multirow{2}{*}{Source}} & \textbf{\multirow{2}{*}{Unit}} & \multicolumn{3}{c|}{\textbf{Dataset}} \\\cline{3-5}
    
      &  & \textbf{The Hill} & \textbf{COVID-19} & \textbf{\textit{Buzzfeed}} \\\hline
     \textit{Breitbart} & Articles & 330 & 35 & 265 \\
     & Sentences & 5,020 & 693 & 3,934 \\\hline
     \textit{Vox} & Articles & 75 & 671 & 62 \\
     & Sentences & 3,561 & 23,089 & 1,637 \\\hline
     \textit{Buzzfeed} & Articles & --- & --- & 30 \\\hline
     \textit{The Hill} & Articles & 147 & 124 & --- \\\hline
     
 \end{tabularx}
\end{table}

\subsection{Text Classification Pipeline}
We decided to demonstrate the \textit{Out-of-Context Summarizer} by altering the context based on the U.S. political spectrum. In order to make the sentences appear right or left leaning, we needed an automated way to tell whether a piece of text is left or right-leaning. To accomplish this and meet the requirements we anticipate that an adversary would encounter in a practical scenario, we wanted a pipeline that could be re-trained for different contexts, performed reasonably well on imbalanced datasets with minimal tuning, and we needed it to provide us with a probability of a sentence being conservative or liberal. We expected that someone trying to use this method would not necessarily have the technical skills required to fully understand how the language model works, so it needed to perform well out of the box. To meet these requirements, our final model pipeline was as follows:

$$
TF-IDF \rightarrow SMOTE \rightarrow Random Forest Classifier
$$

\vspace{0.3cm}

TF-IDF is a text vectorizer that weights each word in a sentence so less common words have a bigger impact on the value used to represent each word. We did customize this step a little bit, to include bigrams and trigrams, as well as to provide the standard list of English stopwords that Scikit-Learn provides. To balance the classes, we used the Synthetic Minority Oversampling TEchnique, or SMOTE. SMOTE generates new vectors to represent sentences in the less dominant class, to make it so that the model has the same number of conservative and liberal sentence vectors. Finally, we used a Random Forest Classifier to actually make the predictions. We selected the random forest classifier \cite{Breiman} because it works relatively reliably with minimal tuning (in fact we got reasonable performance without tuning the model at all), it is relatively fast, and the Scikit-Learn implementation of the RandomForestClassifier has a built in \texttt{.predict\_proba()} method that returns the probability for each prediction \cite{RandomForestClassifier, sklearn}, which is required to weight sentence scores for the text summarization. The model performance metrics are outlined in Table \ref{tab:metrics}. 


\begin{table}[!h]
    \centering
    \caption{\textit{Out-of-Context Summarizer}: Performance metrics calculated on a test set that consists of 10\% of the total dataset size, comparing performance on different datasets without hyperparameter tuning.}
    \label{tab:metrics}
 \begin{tabularx}{\linewidth}{|Y|Y|Y|Y|Y|}
  \hline
     \multicolumn{5}{|c|}{\textbf{The Hill}} \\\hline
 \textbf{Class} & \textbf{Samples} & \textbf{Precision} & \textbf{Recall} & \textbf{Accuracy} \\
  \hline
   \textit{Breitbart} & 503 & 0.94 & 0.94 & --- \\
   \textit{Vox} & 356 & 0.92 & 0.92 & --- \\
   Overall & 859 & 0.93 & 0.93 & 0.93 \\
  \hline
   \multicolumn{5}{|c|}{\textbf{COVID-19}} \\\hline 
    \textbf{Class} & \textbf{Samples} & \textbf{Precision} & \textbf{Recall} & \textbf{Accuracy} \\\hline
   \textit{Breitbart} & 78 & 0.58 & 1.00 & --- \\
   \textit{Vox} & 2301 & 1.00 & 0.99 & --- \\
   Overall & 2379 & 0.79 & 0.99 & 0.99 \\  
   \hline
    \multicolumn{5}{|c|}{\textbf{Buzzfeed}} \\\hline 
    \textbf{Class} & \textbf{Samples} & \textbf{Precision} & \textbf{Recall} & \textbf{Accuracy} \\\hline
  \textit{Breitbart} & 391 & 0.93 & 0.91 & --- \\
\textit{Vox} & 167 & 0.81 & 0.84 & --- \\
 Overall & 558 & 0.87 & 0.88 & 0.89 \\  
   \hline
 \end{tabularx}
\end{table}

Overall, these classifiers performed reasonably well without any special hyperparameter tuning, and all of these models were trained on identical pipelines. Each of the classification models that we set up had a class imbalance, with the most dramatic being the class imbalance with the results for COVID specific articles. While our scraper sampled 23,089 sentences in \textit{Vox} articles discussing COVID-19, it only found 693 \textit{Breitbart} sentences discussing the pandemic. Despite the relative imbalance, the SMOTE oversampling improved the recall of the \textit{Breitbart} sentence identification so all \textit{Breitbart} sentences were identified, though the precision lacked, and only 58\% of those samples were confirmed to be \textit{Breitbart} sentences. For our application, the weak precision, means text summaries for COVID-19 related articles in \textit{The Hill} are more likely to incorrectly include sentences that are liberal in conservative summaries. The success of these models without hyperparameter tuning, means that generating summaries for new sources, or based on new types of data is as simple as retraining the model on different data.

We acknowledge that in addition to the systematic differences between \textit{Vox} and \textit{Breitbart} articles which we attempt to model, the classifiers identify other patterns. These patterns include differences between the style guides that the different news sources use, as well as random noise in the word choices used by different authors or editorial approach. To look inside the model we used the LIME representations \cite{ribeiro}, which we calculated using the \texttt{eli5} python library \cite{eli5}, then we tested the model using a few sentences that we wrote (i.e. they did not appear in the dataset). One of this sentences was: ``Alexandria Ocasio-Cortez's terrible green new deal will destroy our way of life.'' 

Between \textit{Breitbart} and \textit{Vox}, the classifier determined that the sentence had an 87.8\% chance of being in \textit{Breitbart}, based on the permutation importances in Table \ref{tab:predictions}. It looks like the language that makes the text more likely to appear in \textit{Breitbart} is the words based aound the ``way of life,'' and the name ``Alexandria Ocasio-Cortez.'' Additionally, some adjectives like ``terrible'' seem to register as being more typical among the \textit{Breitbart} sentences. This could reflect a difference between the two style guides, where \textit{Vox} may not give writers as much freedom to use adjectives as writers in \textit{Breitbart}, or it could be that the \textit{Vox} editors are more likely to find alternatives for those specific adjectives. It is also possible that the data we scraped happened to only include the word terrible in the text scraped from \textit{Breitbart}, and overall, if we had scraped more data, then the word ``terrible'' would appear as frequently (or even more so) in \textit{Vox} than in \textit{Breitbart}.

\begin{table}[H]
    \centering
    \caption{\textit{Interpreting Classification Predictions}: Permutation importance of the sentence ``Alexandria Ocasio-Cortez’s terrible green new deal will destroy our way of life'' illustrating the model's left- or right-leaning sentence classification.}
    \label{tab:predictions}
 \begin{tabularx}{\linewidth}{|c|Y|c|Y|}
  \hline
   \textbf{Weight} & \textbf{Feature} & \textbf{Weight} & \textbf{Feature} \\\hline
   +0.878 & life & +0.739 & way \\
   +0.287 & new & +0.268 & terrible \\
   +0.187 & Alexandria & +0.166 & ocasio \\
   +0.103 & green & +0.076 & cortez \\
   +0.061 & destroy our & +0.058 & s terrible \\
   +0.037 & Cortez s & +0.030 & will destroy \\
   +0.027 & terrible green & -0.013 & way of \\
   -0.035 & s & -0.063 & our \\
   -0.164 & deal & -0.176 & destroy \\
   -0.489 & <BIAS> & --- & --- \\
  \hline
 \end{tabularx}
\end{table}

\subsection{Text Summarization Implementation}
Our text summarization model is a modification of an extractive text summarization model \cite{Ferreira} to select the sentences that the classifier determines to be the most likely to be from \textit{Vox} or \textit{Breitbart}. 

To make the summaries seem conservative or liberal, we use the probability that a sentence is in a class. The base score that we started out with, and built up from, is a sum of the total appearances of each word in the article for each word in a sentence. To calculate the score for a given sentence $s$, we are adding up the total number of appearances $n$ of each word $w$ in the full article:

$$
score(s)=\sum^{s}_{w=1} n_w
$$

\vspace{0.3cm}

On its own, the probability that the model thinks that a sentence is liberal or conservative doesn't weigh the sentences enough. To overcome this, for $p_1$ the probability that a sentence $s$ is in class $1$, we are taking it to the power $x$:

$$
weighted\ score(s) = p_1^x \sum^{s}_{w=1} n_w
$$

\vspace{0.3cm}

An additional benefit of this parameter is that it allows the adversary to consciously throttle how much of an impact they would like for the machine learning model to have on the sentence summaries, allowing the adversary to use a boiled frog approach and gradually increase how much influence the prediction probabilities might have on the final output. In real articles, some sentences are longer than others. We want to make sure that the sentences that are very long are not considered more important than others, because they are able to include more words. For the length $l$ of a sentence (calculated by the number of words in the sentence), we used a simple binomial expression $lwf$, with upper and lower cutoffs:

$$
    lwf(l)= 
\begin{cases}
    1, & \text{if } al^2+bl+c \geq 1\\
    al^2+bl+c,& \text{if } 0 < al^2+bl+c < 1\\
    0,              & \text{otherwise}
\end{cases}
$$

\vspace{0.6cm}

We wrote this function like this to allow for rapid experimentation with different first and second order functions. So far we have set $a$ as $\frac{-1}{2^{11}}$, $b$ as $0$, and $c$ as $1.1$. This could also be adjusted to even further reduce the length of sentences, or the values could be reduced to allow longer sentences, or dropped entirely. In practice, one benefit of the weight function is to help correct errors with the sentence tokenizer function separating articles into sentences, which occasionally misses a sentence break.

\subsection{Accompanying Hashtag Recommendations}
In the event that adversaries might utilize out-of-context summaries on social media platforms like Twitter or Parler, they may seek to manufacture hashtags to accompany the text and potentially amplify the reach of the content. To this end, \textit{Out-of-Context Summarizer} can also extrapolate potential hashtags regarding a specific article. While a hashtag is too reductive for summarization, it nonetheless can grab a user's attention enough to read a summary accompanying it. Such an approach for crafting politically polarizing Twitter hashtags was used by the Russian troll farms during the US elections in 2016 \cite{Badawy}, for example. To start out with, we calculated the expected number of times that each word would appear in the article and we created a contingency table $P$ where each cell represented the number of times a single word being used in \textit{Breitbart} or \textit{Vox}, with $p_i$ representing the total number of times in the dataset a word was used, and $p_j$ representing the number of words in a specific news source. We used $n$ to represent the total number of occurrences of all words in \textit{Vox} and \textit{Breitbart}. The equation below provides a table of the expected values:

$$E_{i,j}=np_ip_j$$ 

\vspace{0.2cm}

A short list of 15 hashtags is initially complied, based on which words were more likely to make a post seem politically polarizing. To identify words more closely associated as being interpreted as either liberal or conservative, we identified a hashtag score $h$ for a word $w$ that we multiplied the number of times a word appears in an article $a$, by the feature importance $f$ of the feature, by the difference $d$ between the number of times a word appeared in the \textit{Breitbart} or \textit{Vox} dataset and the expected number of times each word would appear:

$$h(w)=ad$$ 

\vspace{0.2cm}

The list of individual words gives an opportunity to leverage fifteen keywords that the adversary can choose between to create hashtags to supplement the text summaries. It is relatively easy to extend the \textit{Out-of-Context Summarizer} to combine the keywords into bigrams or trigrams for more realistic hashtags, but we opted out not to do in order to provide the opportunity for a user to give the ``final touch'' of combining reinforcing hashtags. Another reason is to avoid difficult cognition of the \textit{Out-of-Context Summarizer} output because concatenated text strings in a hashtags are harder to decipher without spaces between words \cite{Pancer}. With that, we wanted to allow for users to have an opportunity for ownership in the adversarial text summarization, or more so, to present our language model as a possible trolling content feeder. We are aware that this could be an inhibiting factor for defenses, because adversaries could use one keyword and then add words outside of the recommendations in an unpredictable manner, and we discuss the human agency in weaponizing adversarial text summarization in our extensive ethical treatise in Section \ref{sec:ethics}.

\section{Out-of-Context Summaries} \label{sec:examples}
In this section we provide example output of the \textit{Out-of-Context Summarizer} from the datasets mentioned above. We chose to show verbose summaries that balance our the lower bound of 280 characters posts imposed on Twitter and the upper bound of 1000 character limit or parleys imposed by Parler. We chose this because one can easily create a conversational model by splitting the verbose summary into two or three consecutive tweets in a thread. 

\subsection{Examples}
\subsubsection{The Hill}
Table \ref{tab:hill_liberal_summary} provides an example liberal-leaning summary of \textit{The Hill's} article on a proposed way ahead after a couple of unusually polarizing and dramatic election cycles \cite{McGee}. The liberal summary discusses the ``heroic efforts'' of election workers, and it discusses the roles that businesses play. It also mentions wealthy individuals, which makes parallels the main premise of the article that funding an election defaults to wealthy individuals if Congress is not willing to step in. The recommended hashtag terms provided a couple fruitful combinations, which appear left-leaning, at face value, in the position against ``big money'':  \#AmericanCrisis, \#NeedAction, \#WorkersSupport. In the context of social media trolling, the summary is 697 characters long and could be split in three posts to create more of a conversational model of trolling \cite{Higashinaka} for Twitter and be taken verbatim to Parler. 


\begin{table}[!h]
\small
    \centering
    \caption{\textit{The Hill}: Liberal Summary}
    \label{tab:hill_liberal_summary}
 \begin{tabularx}{0.9\linewidth}{|X|}
  \hline
\multicolumn{1}{|Y|}{\textbf{Why Congress should provide sustained funding in elections}}  \\\hline
  \textit{  In addition to the heroic efforts of election workers, private philanthropy and businesses stepped up to fill some of the gaps after Congress deadlocked on a bill that would have provided increased emergency funding for states and localities to administer elections. While community leaders have important roles to play in fostering civic engagement, funding elections should not be the responsibility of businesses, nonprofits, or wealthy individuals. Lawmakers from both parties must uphold their Constitutional oaths to support a democracy that guarantees free and fair elections — and they must recognize that free and fair elections depend on both preparation and sufficient, steady funding.} \\
  \hline
  \multicolumn{1}{|Y|}{\textbf{Hashtag Keywords Recommendations}} \\\hline 
  like, new, support, way, need, help, year, community, important, workers, start, crisis, American, address, action \\
  \hline
 \end{tabularx}
\end{table}

The conservative summary of the same article shown in Table \ref{tab:hill_conservative_summary} builds on the roles Congress could have in ensuring safe and secure elections without directly undermining the option for private election funding. The summary ends with a sentence that discusses the ``critical role'' that ``private donors'' play, emphasizing a position of approval for the support coming from wealthy individuals. Though both instances summarize the article fairly effectively, both of the summaries play up one side of the discussion a little bit more heavily. Because all of the sentences were written by a human writer from \textit{The Hill}, neither of the summarizations are conspicuously liberal or conservative. If the summary raised the reader's suspicion, they would find each of the offending sentences in the original article. The recommended conservative-leaning hashtags give some promising ideas, particularly \#ElectionResponsibility, \#PrivateFunding, both of which might help to reinforce a three-piece conversational Tweet or a one-piece Parler post with the 566 characters of content. 

\begin{table}[!t]
\small
    \centering
    \caption{\textit{The Hill}: Conservative Summary}
    \label{tab:hill_conservative_summary}
 \begin{tabularx}{0.9\linewidth}{|X|}
  \hline
\multicolumn{1}{|Y|}{\textbf{Why Congress should provide sustained funding in elections}} \\
  \hline
\textit{Members of Congress, however, take such an oath, and it should be their responsibility to ensure that American elections are both safe and sufficiently funded. The challenges to holding safe and secure elections in 2020 were imminently predictable, and the need for robust federal assistance was clear since last spring — for items like personal protective equipment, poll worker recruitment, and additional equipment to process the record numbers of absentee ballots. This year, private donors played a critical role in ensuring our nation could hold our elections.} \\
  \hline
   \multicolumn{1}{|Y|}{\textbf{Hashtag Keywords Recommendations}} \\\hline 
  2020, election, states, federal, members, private, run, safe, general, congress, responsibility, nation, funding, director \\
  \hline
 \end{tabularx}
\end{table}

\subsubsection{COVID-19}
As trolls usually seek to target a specific polarizing topic or event \cite{nspw2020}, we also tested \textit{Out-of-Context Summarizer} on a topic-specific dataset, consisting of articles related to the COVID-19 pandemic. In order to create conservative and liberal summaries of COVID-related articles on \textit{The Hill}, \textit{Vox}, and \textit{Breitbart}, we utilized each site's respective search functionality (additionally, Vox had a dedicated ``Coronavirus" tab to pull articles from). Tables \ref{tab:covid_liberal_summary} and \ref{tab:covid_conservative_summary} demonstrate the ability of \textit{Out-of-Context Summarizer} to produce the ``tl;dr" of articles related to COVID-19 that could be taken to verbatim or in conversational format to Parler or Twitter.


\begin{table}[h]
\small
    \centering
    \caption{COVID-19: Liberal Summary}
    \label{tab:covid_liberal_summary}
 \begin{tabularx}{0.9\linewidth}{|X|}
  \hline
\multicolumn{1}{|Y|}{\textbf{Progress regarding the passing of COVID-19 relief legislation}} \\
  \hline
\textit{While the Senate made modest changes to the legislation, some of those changes undermined parts of the bill I do support, and others were insufficient to address my concerns with the overall size and scope of the bill, Golden said. But Senate Democrats replaced the GOP amendment hours later with a deal of their own so that the weekly unemployment insurance payments are at \$300 and last through Sept. 6. Another major provision of the bill provides \$1,400 stimulus checks for individuals making \$75,000 or less with phased-out partial payments for those earning up to \$80,000.} \\
  \hline
   \multicolumn{1}{|Y|}{\textbf{Hashtag Keywords Recommendations}} \\\hline 
   pandemic, trump, said, vaccine, going, stimulus, help, testing, republicans, far, democrats, local, unemployment, schools, billion \\
  \hline
 \end{tabularx}
\end{table}

The example article discusses the latest COVID-19 Relief Bill and its intended provisions \cite{COVIDSumm}. Each summary highlights a different part of the article and larger narrative surrounding COVID-19 legislation, in order to take it out of context. In 582 characters, the liberal summary expounds on the specific bill amendments, emphasizing the parts that will directly benefit some constituents, since part of the summary is spoken in the first person tense, it may be more effective for readers. Some of the suggested hashtag terms provided a couple potential combinations that may have the ability to amplify the chosen out-of-context liberal-leaning narrative if posted to social media: \#DemocratsHelp and \#VaccinesHelp might be used to convey a positive message regarding the impact of Democrat-backed COVID-19 relief legislation while conversely, \#RepublicansDontHelp or a hashtag like \#TrumpSaid might be used to mock the opposition party and amplify negative sentiment associated with the GOP (used extensively to ridicule and troll on statements about injecting disinfectants, for example).

With only 424 characters, the conservative-leaning summary developed by \textit{Out-Of-Context Summarizer} seems to highlight the lack of bipartisan cooperation for the bill's passing, in stark contrast to COVID-19 relief measures passed in 2020. The recommended hashtag keywords may not have successfully captured the negative sentiment within the Republican Party associated with the passing of COVID-19 relief legislation but an adversary could identify the negative context and alter hashtags while still using recommended keywords like: \#CorruptDemocrats, \#SenateCronies, or \#BidenAdministrationFailure. A conversational model of tweets carrying alternative narratives was characteristic for the tweeting during the pandemic and both summaries could pass muster in a series of three consecutive Tweets \cite{Chen} or one ``influencer'' Parler post \cite{chi2021}.  

\begin{table}[h]
\small
    \centering
    \caption{COVID-19: Conservative Summary}
    \label{tab:covid_conservative_summary}
 \begin{tabularx}{0.9\linewidth}{|X|}
  \hline
\multicolumn{1}{|Y|}{\textbf{Progress regarding the passing of COVID-19 relief legislation}} \\
  \hline
\textit{House Democrats cleared the legislation by a 220-211 vote on Wednesday, after the Senate passed it in a 50-49 vote on Saturday. By contrast, past pandemic relief measures enacted last year after protracted negotiations between the Democratic-House, GOP Senate and the Trump administration passed with bipartisan support. Only one centrist Democrat, Rep. Jared Golden (Maine), defected from his party during Wednesday’s vote.} \\
  \hline
   \multicolumn{1}{|Y|}{\textbf{Hashtag Keywords Recommendations}} \\\hline 
   Wednesday, house, administration, party, children, democrat, senate, relief, legislation, passed, includes, year, Saturday, golden, popular\\
  \hline
 \end{tabularx}
\end{table}

\subsubsection{BuzzFeed}
In the final iteration of \textit{Out-of-Context Summarizer}, we selected a fourth news source to analyze, \textit{BuzzFeed News}, since it regularly uploads a lot of text content. We summarized a collection of \textit{BuzzFeed} articles, including the article utilized in Tables \ref{tab:buzzfeed_liberal_summary} and \ref{tab:buzzfeed_conservative_summary}, discussing the status of former President Trump's post-election court proceedings to challenge the election outcome \cite{BuzzSumm}. The liberal-leaning summary seems to contextualize the failure by the Trump campaign in court after the election, regarding ``meritless" claims of interference and even, unsympathetic Trump-appointed judges. In 616 characters, the summary also manages to touch on a major point of polarization, suggesting that the Trump campaign used its post-election court proceedings to ``promote lies about widespread voter fraud." If posting the summary to social media, some of the recommended keywords could be used to construct hashtags like \#FakeEvidence, \#RepublicansLost, or \#BidenWon to potentially amplify the post and its out-of-context narrative. 

\begin{table}[htpb]
\small
    \centering
    \caption{BuzzFeed: Liberal Summary}
    \label{tab:buzzfeed_liberal_summary}
 \begin{tabularx}{0.9\linewidth}{|X|}
  \hline
\multicolumn{1}{|Y|}{\textbf{Status of the Trump campaign's post-election litigation}} \\
  \hline
\textit{They petitioned the Supreme Court to hear a small number of these cases, and the justices either rejected them right away or didn’t take any action before Biden was sworn in on Jan. 20, a clear sign that they wouldn’t interfere. Trump and his allies had used postelection legal challenges to promote lies about widespread voter fraud, and they denounced the judicial system, including the Supreme Court, as biased when they repeatedly lost. Judges at every level — including some who were nominated by Trump — concluded that these cases were either procedurally deficient or, after reviewing the evidence, meritless.} \\
  \hline
   \multicolumn{1}{|Y|}{\textbf{Hashtag Keywords Recommendations}} \\\hline 
   filed, right, court, didn, clear, change, republican, evidence, federal, day, small, wasn, won, case, action \\
  \hline
 \end{tabularx}
\end{table}

The lengthier of the two summaries, with 723 characters, the conservative summary is much less harsh regarding the losses, and targeted objective sentences with little opinion and more factual content. \textit{Out-Of-Context Summarizer} seemingly chose less-emotional sentences which were used in the original article solely to convey the facts without offering much commentary since the outcome was unfavorable to the Republican Party. Some of the recommended keywords were already prevalent in hashtags that went viral at the height of the election challenges including \#NotMyPresident, \#IllegitimateElection, or \#WeLovePresidentTrump, suggesting that \textit{Out-Of-Context Summarizer} could be used to anticipate narrative trends on social media in the future. 

\begin{table}[htpb]
\small
    \centering
    \caption{BuzzFeed: Conservative Summary}
    \label{tab:buzzfeed_conservative_summary}
 \begin{tabularx}{0.9\linewidth}{|X|}
  \hline
\multicolumn{1}{|Y|}{\textbf{Status of the Trump campaign's post-election litigation}} \\
  \hline
\textit{Former president Donald Trump has officially lost all of his postelection legal challenges in the US Supreme Court, after the court announced Monday that the justices had refused to take up his final case out of Wisconsin. Trump and his Republican allies lost more than 60 lawsuits in state and federal courts challenging President Joe Biden’s wins in Wisconsin and a handful of other key states. A handful of cases remained active even after Biden took office, and on Feb. 22 the Supreme Court issued a round of orders officially refusing to hear cases that Trump had brought in Pennsylvania and Wisconsin (he filed multiple cases in that state), as well as other Republican-backed challenges in Pennsylvania and Michigan.} \\
  \hline
   \multicolumn{1}{|Y|}{\textbf{Hashtag Keywords Recommendations}} \\\hline 
  biden, president, state, joe, house, white, states, trump, monday, donald, legal, office, election, march, pennsylvania \\
  \hline
 \end{tabularx}
\end{table}

\subsection {Linguistic Analysis}
For each iteration of \textit{Out-of-Context Summarizer}, we used the Linguistic Inquiry and Word Count (LIWC) tool to perform a quantitative content analysis of the three summarization datasets (i.e., all compiled original, conservative, and liberal summaries) \cite{Pennebaker}. Original summaries were not manipulated to appear right or left-leaning, but \textit{Out-Of-Context Summarizer} was trained on text from \textit{Vox} and \textit{Breitbart}, to select liberal and conservative-leaning summaries. LIWC has been extensively used to study the politically polarizing rhetoric on wide range of topics with datasets similar to ours containing news articles \cite{Graham, Sterling, Wojcik}. We decided to compare the summarization datasets on the four main categories: (1) analytical thinking; (2) clout; (3) authenticity; and (4) emotional tone. We didn't analyzed each article's three summaries individually, as LIWC requires a larger-scale database to perform effectively. 
 
 
 \subsubsection{The Hill}
We first performed a linguistic analysis on the initial model of \textit{Out-Of-Context Summarizer} in which summaries were developed from articles published by \textit{The Hill}. Table \ref{tab:hill_liwc} illustrates the results of the LIWC analysis on \textit{The Hill} dataset. The high ``analytical thinking'' score for all three summary sets indicate that overall, they follow logical and thoughtful patterns, although this is not highly surprising considering that all summarizations include unaltered sentences from legitimate news articles published by \textit{The Hill}. Potentially notable is the decreased analytical score of liberal summaries compared to both the unaltered summaries and conservative groups which are fairly similar. 

All summaries are rated similarly in the``clout" category, for perhaps a similar reason: this metric measures confidence level, and journalists may want to exude a high level of confidence in their writing, in hopes of adding a perception of legitimacy to their work. In the ``authenticity" category, the liberal summaries contain the most unique content, but all three groups fall under 50, indicating that the summaries are prone to regurgitate known phrases and narratives. And finally, the ``emotional tone metric indicates that while liberal summaries picked slightly more positive sentences than the original group, the conservative summaries chosen were more negative. 

\begin{table}[!h]
\caption{LIWC Scores of \textit{The Hill} Summaries}
\label{tab:hill_liwc}
\centering
\begin{tabularx}{\linewidth}{|Y|Y|Y|Y|}
\hline
\textbf{Analytic Thinking} & \textbf{Clout} & \textbf{Authenticity} & \textbf{Emotional Tone} \\\hline
\multicolumn{4}{|c|}{\textbf{Original Summaries}} \\\hline
97.31 & 71.07 & 21.20 & 49.00  \\\hline
\multicolumn{4}{|c|}{\textbf{Liberal Summaries}} \\\hline
94.50 & 73.34 & 22.12 & 49.62	 \\\hline
\multicolumn{4}{|c|}{\textbf{Conservative Summaries}} \\\hline
97.87 & 72.70 & 20.34 & 46.34 \\\hline
\end{tabularx}
\end{table}

 \subsubsection{COVID-19}
After the initial iteration of \textit{Out-Of-Context Summarizer}, we scraped additional text from \textit{The Hill}, \textit{Vox}, and \textit{Breitbart} related to the ongoing COVID-19 pandemic and performed a subsequent linguistic analysis of accumulated summary groups (i.e. unaltered original summaries, liberal-leaning summaries, conservative summaries). Consistent with the LIWC results of the previous \textit{The Hill} iteration, each group of COVID-19 summaries ranks fairly high in the category of ``analytical thinking", with liberal summaries ranking lowest of the three summary groups in both iterations. As shown in Table \ref{tab:covid_liwc}, all summary groups score similarly in the ``clout" category, indicating that each maintains a fairly confident writing style, which is also unsurprising considering the source of the text. 

In the ``authenticity" category, conservative summaries include seemingly more unique content than the original and liberal summaries; conversely, the liberal summaries contain even less unique content than the original group. Finally, each group of summaries has an ``emotional tone" score under 50, which indicates a negative and impolite style of writing, with the conservative group of summaries falling even lower in this category. Liberal summaries selected more positive sentences, increasing the ``tone" score. This suggests that while both groups of summaries still selected negative sentences, conservative sentences preserved negative content from the original articles, potentially increasing the likelihood of polarization of a topic like COVID-19 if viewed out of context. 

\begin{table}[!h]
\caption{LIWC Scores of COVID-19 Summaries}
\label{tab:covid_liwc}
\centering
\begin{tabularx}{\linewidth}{|Y|Y|Y|Y|}
\hline
\textbf{Analytic Thinking} & \textbf{Clout} & \textbf{Authenticity} & \textbf{Emotional Tone} \\\hline
\multicolumn{4}{|c|}{\textbf{Original Summaries}} \\\hline
97.30 & 69.88 & 23.54 & 41.06  \\\hline
\multicolumn{4}{|c|}{\textbf{Liberal Summaries}} \\\hline
96.89 & 69.56 & 21.45 & 47.83	 \\\hline
\multicolumn{4}{|c|}{\textbf{Conservative Summaries}} \\\hline
97.47 & 69.35 & 28.17 & 41.76 \\\hline
\end{tabularx}
\end{table}


\subsubsection{BuzzFeed}
In our final iteration of \textit{Out-Of-Context Summarizer}, we targeted an additional news source, \textit{BuzzFeed}, and scraped news articles, then summarized the article three different times (i.e. unaltered original, left-leaning liberal, right-leaning conservative) and performed a linguistic analysis on each compiled set of summaries, shown in Table \ref{tab:buzzfeed_liwc}. The trend established in the previous two iterations regarding the ``analytic thinking" category was present in this dataset as well, with original and conservative summaries seemingly following more logical thought-patterns than the liberal group of summaries. On average, the \textit{BuzzFeed} dataset had the highest ``clout" scores of the three iterations, suggesting the high confidence of writers at \textit{BuzzFeed}.

\begin{table}[!h]
\caption{LIWC Scores of \textit{BuzzFeed} Summaries}
\label{tab:buzzfeed_liwc}
\centering
\begin{tabularx}{\linewidth}{|Y|Y|Y|Y|}
\hline
\textbf{Analytic Thinking} & \textbf{Clout} & \textbf{Authenticity} & \textbf{Emotional Tone} \\\hline
\multicolumn{4}{|c|}{\textbf{Original Summaries}} \\\hline
96.19 & 79.35 & 10.54 & 27.48  \\\hline
\multicolumn{4}{|c|}{\textbf{Liberal Summaries}} \\\hline
92.32 & 75.91 & 14.49 & 19.52	 \\\hline
\multicolumn{4}{|c|}{\textbf{Conservative Summaries}} \\\hline
96.46 & 79.83 & 13.37 & 22.27 \\\hline
\end{tabularx}
\end{table}

Of the three groups of summaries, the liberal set scored the highest in ``authenticity" suggesting those summaries contained the most unique content. Additionally, in regard to ``authenticity" of summary contents from \textit{BuzzFeed} articles, on average, it contained the least original, unique content of the three datasets we analyzed using LIWC. Potentially most striking is the ``emotional tone" category, in which the original summaries scored significantly higher, suggesting a negative tone, but not nearly as negative as either the liberal or conservative summaries, which both seemingly targeted more gloomy sentences from \textit{BuzzFeed} articles. The tendency of \textit{Out-Of-Context Summarizer} to select negative or impolite sentences may lead to increased polarization and by extension, a more favorable outcome for potential adversaries. 

\section{Guarding Against Adversarial Language Modeling}
Our implementation of \textit{Out-of-Context Summarizer} worked easily because we were able to very quickly find a high-quality dataset, which fit our purposes. We targeted content that was easily obtainable, and was homogeneous relative to our other sources, both in terms of its lexicon and the contexts in which the three sources were written, that allowed the machine learning models to effectively generalize to the neutral news source. We selected the news sources because they were especially convenient to access. None of the news sources used were behind a paywall, nor did they require a login, and we did not encounter any defenses to prevent us from using the same IP address to systematically access the websites directly in the order they appear in main pages. Additionally, the language is relatively easy to parse using the \texttt{bs4} \cite{bs4} python library, which enabled rapid prototyping and scraping of data. We selected these news sources because all of them are written in a relatively similar context. One news-source that we initially considered for our centrist news source was the \textit{BBC}. We decided against the \textit{BBC} articles because they are written using British spelling, so the sentence classification would not be as simple. Additionally, articles published in the \textit{BBC} are written for an international audience. They do not make the same assumptions or provide details without explanations to the same extent that newsletters tailored toward American audiences might. 

Protecting against an adversarial implementation of a tool like \textit{Out-of-Context Summarizer} may not be simple, as it weaponizes legitimate news sources in order to change the context of content, and not the content itself. Potentially one of the best defenses to text summarization being taken out of context is read a piece in its entirety to alleviate the constraints of any tool's inadvertent, or intentional, biases. Yet this is easier said than done. Recently, Twitter tested a means of soft moderation, in hopes of confronting this same issue we now face. Select Twitter users were prompted with a message when trying to retweet a post with an attached article link: ``Headlines don't tell the fully story. You can read the article on Twitter before retweeting'' \cite{TwitterUpdate2}. More research must be done into the effectiveness of soft moderation tactics, and warnings targeted at specific content may increase the likelihood of the ``backlash'' effect when users double-down of previously-held beliefs because of the warning \cite{Pennycook}. 

Researchers in a preliminary study found that some means of soft moderation on social media sites like Twitter are more promising than others, including warnings which fully obscure content like Twitter's ``Read First'' warning \cite{soups2021}. Levying soft moderation techniques like these may encourage users to read the full context of news content, as well as serve as a reminder of the dangers of out-of-context content in perpetuating the spread of alt-narratives online. For sites like Parler, which prides its brand on the lack of moderation, containing the spread of alt-narratives propagated by humans or machines may deem to be more difficult, as the site is designed to help alt-narratives flourish \cite{chi2021}.

\section{An Adversarial Disclosure Test} \label{sec:ethics}
\subsection{Ethical Argumentation}
The misuse of adversarial AI research, particularly automated language models, appears to be a complex problem that cannot be solved with the boilerplate ethical argument for responsible disclosure of security vulnerabilities \cite{Shevlane}. The ``security value'' of disclosing a vulnerability outweighs the risk of an adversary exploiting it, the argument goes, because the patched system and the improved risk management will make for a harder target in the future. Disclosing a fake text language model might not outweigh the risks of an adversary exploiting it, say for trolling or spreading misleading text summarizations. First, there is no obvious patch because we live in a post-truth society where fake text can serve as truth as long as is sufficiently contrarian to positions that one dislikes \cite{Wood}. A defender could prevent buffer overflow, but could hardly prevent an overflow of automated text summaries because the summary could be used as a seed for humans to generate and evolve new alternative narratives \cite{Stewart}. Second, managing the risk of trolling or spreading  alternative narratives is a thorny issue inhibited with the constitutional right for free speech. Take for example the effort for soft moderation on social media platforms to curb misinformation \cite{Clayton}. Adding warning labels on seemingly fake text, even if human generated, not just alienated users to abandon mainstream platforms like Twitter and switch to alternatives like Parler \cite{Caulfield}, but inspired some users' further belief in the alternative narratives promoted with the fake text \cite{Nyhan}. 

The intention to publish automated language models despite the adversarial output, after all, is to increase defenders' understanding of the threat from text generation, and contribute to building tools that could guard against the harms \cite{Radford, Zellers, Keskar}. The Kerckhoffs' doctrine that ``security through obscurity doesn't work'' is thus thrown into the mix as an argumentation for publishing fake text generators because an adversary would figure out the language model themselves, if they haven't already, and proceed with exploitation (akin to zero-day vulnerabilities) \cite{eurosp2020}. Adversaries, in the traditional security sense, pose extensive knowledge of targeted systems that help them work through the obscurity or find a zero-day vulnerability. There is no doubt that they could craft a fake text generator, especially after the the revelation of massive human troll farms that generated fake news during the 2020 U.S. Elections \cite{Twitter}. Individual users with virtually no knowledge of adversarial language modeling could become ``adversaries'', by propagating alternative narratives by leveraging existing tools like GPT-2 or Grover. Adversaries could also use rudimentary text modeling, using tools like the TrollHunter[Evader] that uses a Markov chain to subversively replace target words with replacements in trolling Tweets to evade automated trolling detection \cite{nspw2020}. A user needs not to replicate TrollHunter[Evader] to employ the evasion idea when generating their own content. 

The ``vulnerability'' revealed by publishing automated language models, then, has to be guarded on two fronts, not just one - the system patching - as in the traditional system security sense. Defenders, initially, need automated counter models to detect fake text generated by automated language models, but they also need assistance when dealing with fake text generated by humans. On the first account, the battle of the machines is underway, and work already exists to detect nefarious text generation for adversarial purposes \cite{Zellers}. On the second account, the machine-human battle is a bit more complicated. Back to the case of soft moderation on Twitter, evidence shows that automated soft moderation for COVID-19 misinformation tweets is prone to mislabeling simply because targets COVID-19 rumor words/hashtags like ``oxygen'' and ``frequency'' \cite{Zannettou}. Defenders also have to work on the human front to help humans distinguish between fake text generated by a machine and fake text generated by another human. This guard might be even superseded by simply guarding the truth regardless of the text origins, but building such a tool seems quite a Sisyphean task. The fake text as a form of deception is conductive to formation of echo chambers \cite{Himelboim}, conspiracy theories \cite{Starbird}, and even entire social media platforms like Parler \cite{chi2021}, which are hard to be eradicated or dispelled in the era of post-truth with any constructive argumentation \cite{Pennycook}. One could likely expect a similar effect from fake text generated from automated language models. 

The ``adversary knows the system'' might upend the argument for publication of adversarial automated language models but the (dis)balance of offense-defense argumentation, in our view, should be upended by the notion that humans are both the adversaries and the direct victims of fake text. Even before fake generators existed, fake or ill-generated text by humans caused great harm. Take for example the translation of intercepted North Vietnamese messages in the Gulf of Tonkin incident, which replaced the words ``comrades'' with ``boats'' to create an impression of an attack that was sufficient to justify a declaration of war by the Johnson administration \cite{nsa}. No fake text generators, or translators existed in this case, but an adverse effect nonetheless was produced by humans and targeted at humans. Certainly, a publication of a tool for fake text generation or translation could be misused to cause similar harms, but for that to happen, an adversarial intention by a human should probably exist in the first place.   

\subsection{Framework for Weighing Security Costs and Benefits of Disclosure}
\subsubsection{Attacking}
With the argumentation in mind, we apply the framework proposed in \cite{Shevlane} for weighing security costs and benefits of disclosure for our tool \textit{Out-of-Context Summarizer}. First, we consider factors that affect the adversaries' capacity to cause harm. The first factor is \textbf{\textit{counterfactual possession}} or the possibility that the would-be adversary would either independently discover the language modeling behind the \textit{Out-of-Context Summarizer} or learn about it from other adversaries. We believe that it is relatively easy for an adversary to independently discover the language modeling; first, the design described in section \hyperref[sec:ats]{Section 2} is based on well-known language modeling techniques like TD-IDF and SMOTE. Even if we did not publish this paper or the adversary doesn't look into academic papers, they have the option to skim numerous trending blogs of language modeling and find out how these techniques work. The politically polarizing summarization and generation of hashtags is just the flavor we chose to adapt these models into, emphasizing the long-played editorial practice of taking statements out-of-context (also known as ``contextomy'') \cite{McGlone}. This practice of reporting emerged way before any fake text generators surfaced in academic journals or conference proceedings and resourceful adversaries used it to coordinate humans to generate fake text. Take for example the conservative politicians’ quotation of Rev. Martin Luther King in their campaigns to eliminate affirmative action programs in the US \cite{Holmes}. Or the infamous Russian Trolls that took a citation to Hillary Clinton's ``mentally ill'' out of context to suggest that she incites violence on Trump rallies through the Tennessee GOP Twitter account \cite{Twitter}. The \textit{Out-of-Context Summarizer} does a distantly related adversarial text summarization in a rather benign fashion because it only borrows from, but it does not make fake replacements in already published articles. 

The second factor is \textbf{\textit{absorption and application capacity}} of the adversary or the extent to which they are able to grasp and utilize the full potential of \textit{Out-of-Context Summarizer}. The barrier for absorption is low because the adversary needs only to ``get the idea'' of  \textit{Out-of-Context Summarizer} and perhaps even implement it with other language modeling techniques without the need to replicate our work \cite{Radford, Zellers, Keskar}. There is little cost in any path of adaptation. The application capacity could be a bit problematic for \textit{Out-of-Context Summarizer}, but that's the case for all adversarial or fake text generators. An adversary could select biased news articles or even fake text from GTP-2 or Grover for the training phase and summarize texts with hashtags to assign a meaningful context to otherwise coherent-only content. An adversary could pair the \textit{Out-of-Context Summarizer} output with the techniques from the TrollHunter[Evader] to combine a hashtag and trolling content that can evade both soft and hard moderation on Twitter on any topic of their choice, not just COVID-19 \cite{nspw2020}. Take for example the out-of-context summarization and hashtags from the Centers from Disease Control (CDC) about a potential allergic reaction to the first dose of the COVID-19 vaccine, shown in Table \ref{tab:cdc} ~\cite{CDC}: 

\begin{table}[!h]
\small
    \centering
    \caption{CDC Summary}
    \label{tab:cdc}
 \begin{tabularx}{0.9\linewidth}{|X|}
  \hline
\multicolumn{1}{|Y|}{\textbf{What to Do if You Have an Allergic Reaction After Getting A COVID-19 Vaccine}} \\
  \hline
\textit{If you had a severe allergic reaction -- also known as anaphylaxis -- after getting the first shot of a COVID-19 vaccine, CDC recommends that you not get a second shot of that vaccine. If it is not feasible to adhere to the recommended interval and a delay in vaccination is unavoidable, the second dose of Pfizer-BioNTech and Moderna COVID-19 vaccines may be administered up to 6 weeks (42 days) after the first dose. } \\
  \hline
   \multicolumn{1}{|Y|}{\textbf{Hashtag Recommendations}} \\\hline 
   \#delayvaccination, \#analylaxis, \#COVID-19 \\
  \hline
 \end{tabularx}
\end{table}

Using the TrollHunter[Evader], an adversary can manipulate the out-of-context summarization to read more coherently by replacing ``interval'' with ``abstination'' or replace the \#COVID-19 with \#COVIDIOT and use the resulting output as an anti-vaccination alternative narrative for the COVID-19 vaccine. Per the ``Goldilock zone'' reasoning provided in \cite{Shevlane}, the \textit{Out-of-Context Summarizer} fits into the ``script kidde'' case, given that it requires fairly little capacity to absorb the inner workings of the language modelling, but the application range is quite large (as mentioned in the argumentation section above, an adversary could be anybody who's intent is to communicate fake or manipulated text). 

\subsubsection{Defending}
Next, we consider factors that affect the the defenders' ability to mitigate the potential harms. We extend the framework to consider both automated means of defense by also the human ability to discern fake generated text or text generated for adversarial purpose, because we argued that the humans, or the text users, are both the adversaries and the defenders. For the first factor, \textbf{\textit{counterfactual possession}}, we provide a defensive discussion on potential ways to detect an output of the \textit{Out-of-Context Summarizer} with automatic means. We also believe the detection mechanism provided in \cite{Zellers} could be adapted for defensive purposes of the summarized text. A correspondence analysis as proposed in \cite{iwspa2021} could possibly help in detecting the automatically generated polarizing hashtags. On the human front, we don't believe that the \textit{Out-of-Context Summarizer} sounds the alarm on adversarial text summarization, given the debate over the fake text generation surrounding GPT-2 \cite{Shevlane}, the well known existence of ``contextomy'' \cite{McGlone}, or the existence of ```belief echoes''  \cite{Pennycook1}. The preliminary effort of ``defeneding'' from fake or adversarial text with both soft and hard moderation on social media, for example, is out to a rocky start, given that the warnings labels can backfire \cite{Clayton}. But, Twitter recently changed the defense strategy to include strikes \cite{TwitterUpdate} before banning someone's account for sharing adversarial text and experimented with crowd ``fact checkers'' in their Birdwatch program \cite{Ortutay2021}. How these mechanisms will affect the human defense remains to be explored and we believe they will provide the similar aid against the \textit{Out-of-Context Summarizer} output.

For the second factor, \textbf{\textit{absorption and application capacity}}, we believe that the barrier for absorption is low because the defenders, too, need only to ``get the idea'' of  \textit{Out-of-Context Summarizer} and perhaps implement a baseline context checks to detect any drift in the context or the summarized text with both numerical or linguistic analysis as we propose in our defense section. Perhaps same for humans, that could possibly manually go through the input articles and compare the \textit{Out-of-Context Summarizer} summarization, provided they are ready to engage in such a tedious task. Anticipating the applications of \textit{Out-of-Context Summarizer} might seem like a complex task for both the automated and human defenders, given that the adversarial creativity usually stays one step beyond. Not to be discouraged, because this creativity feeds on the contemporary polarizing events --- alternative narratives were generated about the Boston Marathon bombings, the Pope endorsement of the Donald Trump, but also for the human manufacturing of the COVID-19 pandemic and the adverse effects of the vaccine \cite{Starbird, Pieroni}. In both cases, the \textbf{\textit{resources for solution finding}} will be considerable, given the perpetuating adversarial value of adversarial language models for any current or future topic of content. Even if the solutions are available, or could be made available relatively fast, the \textbf{\textit{solutions' effectiveness}} and the \textbf{\textit{solutions' adoption}}, especially by the human defenders, could range widely based on the wide options for application of the \textit{Out-of-Context Summarizer}. 
 
\subsection{The Security Value of Disclosure}
We applied the framework to determine the security value of disclosure of our \textit{Out-of-Context Summarizer} language model. We decided to disclose the idea of adversarial out-of-context text summarization and polarizing hashtag generator with the tool we developed to test such automation. We are aware that the \textit{Out-of-Context Summarizer} might have a slight ``offensive bias'' particularly in combining the generated summaries and polarizing hashtags to perpetuate a contrarian position to the mainstream stance on COVID-19, elections, or any highly contested topic. The security value of disclosure, we believe, lays in appending the argumentation for ``patching social vulnerabilites'' as pointed out in \cite{Shevlane}, but also in highlighting the ease with which an adversary could automate and weaponize the human need for ``contextomy'' to either up the ante on polarization or blackbox probing of defensive mechanisms.

On social media platforms with a imbalanced number of ``influencers'' and ``followers'' like Parler, an ``influencer'' could use the \textit{Out-of-Context Summarizer} to further reinforce a negative sentiment as shown for the case of the notorious conspiracy theory QAnon in \cite{chi2021}. An ``influencer'' could only use the language modeling as an aid, a seed idea to craft an alternative narrative in a way to avoid soft moderation, birdwatching, or strikes on mainstreams platforms like Twitter too. The ``offensive bias'' might not come from a single post, but more so as a blackbox probing of Twitter's algorithms or policies by an adversary for which they can use the \textit{Out-of-Context Summarizer} to generate a vast amount of adversarial text samples accompanied by polarizing hashtags. This comes in handy to append the practice of sharing links by buffering the access to the full content of the articles while capitalizing on their credibility or assumed position on a polarizing issue. The disclosure of our language model, in these regards, serves as an information about a new threat of not just another fake text, but original text automatically taken out of context and possibly appended with polarizing hashtags. It helps the defense, automated or at least the human ''birdwatchers,'' to conceive an automated text summarization beyond the one-dimensional distributing of fakes and originals, but includes a dimension where original text could be taken out of context by automatic means.





\section{Discussion} \label{sec:discussion} 

\subsection{Related Work}
The techniques that we used for text-summarization in this paper, with the exception of the use of the classifier, are covered in detail in a paper by Ferreira et al. \cite{Ferreira}. Text summarization has also been applied in adversarial contexts, with two examples being the 2008 application of extractive text-summarization as a steganography approach to disguise messages \cite{Desoky}. This approach is similar to ours in that it uses text-summarization as a tool for deceiving people into thinking a summarization is sharing one message, but this approach is not intended to actively persuade the target to mis-perceive the manipulated text. Another paper highlighted a possible application of text summarization to help summarize lots of text data about a rival company to build a concise competitor profile \cite{Chakraborti}. This paper uses text summarization to quickly perform a lot of research on a target than might otherwise be easily available. 

Adversarial language manipulation is an active area of research, with a variety of recent developments. One recent example is the python library \texttt{TextAttack}, which provides adversaries with the ability to reliably and precisely change the meaning of text based on a variety of different models \cite{Morris}. Our \textit{Out-of-Context Summarizer} does not need to manipulate words within sentences, because it uses selective summarization to show one perspective within an article, and presents subsets of the article to show a biased perspective of the article. Unlike \texttt{TextAttack}, which manipulates the language of a piece of text, meaning that a targeted reader could possibly see differences between the original and the manipulated text with a closer inspection of the text out.

\subsection{Future Enhancements}
There are many optimizations available that could be built into our text summarizations to improve the substitutions' quality. Our implementation used a minimally viable implementation based on word frequency with optional sentence length optimizations. Possible enhancements could include the use of the TF-IDF statistic, lexical similarity of terms, text case, part of speech for word scores, cue phrases, sentence position, sentence centrality, enhancement using numerical data, bushy path of the node, and aggregate similarity \cite{Ferreira}. Our text summarization mechanism has a lot of different parameters to determine. The ideal parameters might be different for different people, choice of article, and tuning for the application of taking narratives out-of-context. By using a machine-learning driven approach, such as a multi-armed bandit, we might be able to define the text summary parameters based on how a user reacts to the text summaries. Another option would be to automate even more of the process. While an adversary might extend a couple of modules that we built for the \textit{Out-of-Context Summarizer} summaries to post on social media platforms without any human intervention, with ethical considerations, we could develop automatically-generated simulated tweets or parleys for experimentation to gauge user reactions in a lab setting, similar to user studies investigating the effects of soft moderation on Twitter \cite{soups2021}.

\subsection{Abstractive Adversarial Summarization}
In addition to the default \textit{extracting} summaries, \textit{Out-of-Context Summarizer} could be adapted to generate \textit{abstractive} summaries with little effort. We already provided a hint to an adversarial abstractive adaptation when discussing the manipulation of the summary shown in Table \ref{tab:cdc}. The first thing an adversary could do is replace words or hashtags, like we exemplified with the replacement of ``interval'' with ``abstination'' or \#COVID-19 with \#COVIDIOT. The next thing what the adversary could do is borrow sentences from other CDC articles and abstract the severity of the said allergic reaction to the COVID-19 vaccine as shown in Table \ref{tab:cdc_abstractive} \cite{cdcseverity}. Compared to the summary in \ref{tab:cdc}, the abstractive summary adds an additional dimension to the out-of-context summarization for full realization of alternative narratives dissimination by taking out-of-context an entire set of articles on a topic, not just one. 

Though not in an automated fashion, a similar attempt of out-of-context trolling was used by Robert F. Kennedy Jr. in his attempt to discredit the COVID-19 vaccination (which earned him a ban from Instagram) \cite{Chappell}. In this context, the misperception-inducing approach of word manipulation from \cite{eurosp2020}, an adversary could also go further, for example, to append the first summary sentence with \textit{..., which resulted in death in several cases in Florida} in order to hint an extreme outcome of the vaccination. Certainly this moves the summary into the ``fake news" category, and we provided this example to highlight the dangers of abstractive re-purposing of \textit{Out-of-Context Summarizer}. We condemn such use, of course, but insinuations and implausible interpretations of a set of individual facts was, and still is, the essence of trolling on Twitter and especially Parler \cite{chi2021}.  

\begin{table}[!h]
\small
    \centering
    \caption{Abstractive CDC Summary}
    \label{tab:cdc_abstractive}
 \begin{tabularx}{0.9\linewidth}{|X|}
  \hline
\multicolumn{1}{|Y|}{\textbf{What to Do if You Have an Allergic Reaction After Getting A COVID-19 Vaccine}} \\
  \hline
\textit{A severe allergic reaction -- also known as anaphylaxis -- happens within 4 hours after getting vaccinated with the the first shot of a COVID-19 vaccine and could include symptoms such as hives, swelling, and wheezing (respiratory distress). CDC recommends that you not get a second shot of that vaccine. You need to be treated with epinephrine or EpiPen or go to the hospital.} \\
  \hline
   \multicolumn{1}{|Y|}{\textbf{Hashtag Recommendations}} \\\hline 
   \#severereaction, \#analylaxis, \#COVID-19 \\
  \hline
 \end{tabularx}
\end{table}

\subsection{Summary Abbreviation and Expansion}
Both the expressive and abstractive summarizations could also be appended or utilized to generate headlines, a one-sentence summary, or an extended full summary. It's trivial to limit the summary and select one sentence to be a headline, for example, selecting the first sentence in \ref{tab:cdc} verbatim or modifying it to read \textit{CDC recommends that you not get a second shot of that vaccine, if you had a severe reaction after getting the first shot of a COVID-19 vaccine}. An adversary could use the approach for modifying emails from \cite{eurosp2020} to simply make an email appear with a subject line \textit{CDC recommends that you not get a second shot of that vaccine}. Similarly, one could use the approach from \cite{malexa2021} and create a third-party COVID-19 skill that will deliver headlines to Amazon Alexa users by summarizing articles from the CDC website. A similar adversarial twist with Alexa was successful in reducing the perceived accuracy of information as to who gets the vaccine first, vaccine testing, and the side effects of the vaccine, as shown in \cite{dimwa2021}. A full expressive summary could also be generated for the purpose of updating or creating Wikipedia articles with the latest developments of the pandemic, for example. Again, using the approach for evading detection of Wikipedia vandalism \cite{acsac2020}, an adversary could simply decide to minimize mentions of a target vaccine maker, for example, Sinopharm, if the adversary wants to implicitly promote other vaccines. Such a trolling attempt already surfaced on Twitter, promoting homegrown Russian vaccines and undercutting rivals ~\cite{Frenkel}.

\subsection{Limitations}
\textit{Out-of-Context Summarizer}, like every automated text summarization mechanism, comes with a set of limitations. First, the out-of-context summaries are generated using a relatively simple design of $TF-IDF \rightarrow SMOTE \rightarrow Random Forest Classifier$. Other algorithms for text summarization might yield out-of-context summaries different than the ones generated by \textit{Out-of-Context Summarizer}. We also allow for simplistic yet flexible calculation of the weights and sentence tockenization, which are couple of segments of the language model that could also be modified to generate different output. One could potentially choose another approach for word count and calculating the hashtag score to produce a non-overlapping set of hashtag keywords with the set of keywords generated by \textit{Out-of-Context Summarizer}. Second, we deliberately selected four news outlets other than the main agenda setters or any alternative outlets on the far ends of the spectrum. One could try to train \textit{Out-of-Context Summarizer} on other outlets than ours, and, with a high probability, generate a separate liberal or conservative context. A similar conclusion holds for the selection of the articles reported in the time frame we conducted the study - COVID-19, elections, or stimulus as polarizing topics might morph and be incorporated in the trolling agenda in unpredictable ways. Finally, we used a modest infrastructure to implement and test the \textit{Out-of-Context Summarizer} with satisfactory results of generating a summary and recommended hashtag keywords in less than few seconds. Summarizing larger datasets could affect this performance.

\section{Conclusion}
We used the adversarial language model proposed in this paper to generate two versions of the ``tl;dr" conclusion. We hope that the conclusion demonstration together with the argument put forward in the paper can serve the security community to better deal with adversarial language modeling in future. 

\subsection{Liberal-leaning Conclusion}
The summary output of \textit{Out-of-Context Summarizer} algorithm is both \textit{indicative}, alerting the user about the source content, and specifically \textit{informative}, outlining the main contents of the original text (though with an original twist, which makes the \textit{Out-of-Context Summarizer} appealing for adversarial text summarization). For the language modeling, we must note, an adversary might not be using news articles per se; they can train the \textit{Out-of-Context Summarizer} on the textual portion of available Twitter information operations datasets \cite{Twitter}, for example on a particular topic like COVID-19 \cite{nspw2020}. In addition to the default \textit{extracting} summaries, \textit{Out-of-Context Summarizer} could be adapted to generate \textit{abstractive} summaries with little effort. 

\subsection{Conservative-leaning Conclusion}
\textit{Out-Of-Context Summarizer} seemingly chose less-emotional sentences which were used in the original article solely to convey the facts without offering much commentary since the outcome was unfavorable to the Republican Party. For each iteration of \textit{Out-of-Context Summarizer}, we used the Linguistic Inquiry and Word Count (LIWC) tool to perform a quantitative content analysis of the three summarization datasets (i.e., all compiled original, conservative, and liberal summaries) \cite{Pennebaker}. Original summaries were not manipulated to appear right or left-leaning, but \textit{Out-Of-Context Summarizer} was trained on text from \textit{Vox} and \textit{Breitbart}, to select liberal and conservative-leaning summaries.

\bibliographystyle{ACM-Reference-Format}
\bibliography{\jobname}


\begin{thebibliography}{65}


\ifx \showCODEN    \undefined \def \showCODEN     #1{\unskip}     \fi
\ifx \showDOI      \undefined \def \showDOI       #1{#1}\fi
\ifx \showISBNx    \undefined \def \showISBNx     #1{\unskip}     \fi
\ifx \showISBNxiii \undefined \def \showISBNxiii  #1{\unskip}     \fi
\ifx \showISSN     \undefined \def \showISSN      #1{\unskip}     \fi
\ifx \showLCCN     \undefined \def \showLCCN      #1{\unskip}     \fi
\ifx \shownote     \undefined \def \shownote      #1{#1}          \fi
\ifx \showarticletitle \undefined \def \showarticletitle #1{#1}   \fi
\ifx \showURL      \undefined \def \showURL       {\relax}        \fi
\providecommand\bibfield[2]{#2}
\providecommand\bibinfo[2]{#2}
\providecommand\natexlab[1]{#1}
\providecommand\showeprint[2][]{arXiv:#2}

\bibitem[\protect\citeauthoryear{{Badawy}, {Ferrara}, and {Lerman}}{{Badawy}
  et~al\mbox{.}}{2018}]%
        {Badawy}
\bibfield{author}{\bibinfo{person}{A. {Badawy}}, \bibinfo{person}{E.
  {Ferrara}}, {and} \bibinfo{person}{K. {Lerman}}.}
  \bibinfo{year}{2018}\natexlab{}.
\newblock \showarticletitle{Analyzing the Digital Traces of Political
  Manipulation: The 2016 Russian Interference Twitter Campaign}. In
  \bibinfo{booktitle}{\emph{2018 IEEE/ACM International Conference on Advances
  in Social Networks Analysis and Mining (ASONAM)}}. \bibinfo{pages}{258--265}.
\newblock
\urldef\tempurl%
\url{https://doi.org/10.1109/ASONAM.2018.8508646}
\showDOI{\tempurl}


\bibitem[\protect\citeauthoryear{Boukes}{Boukes}{2019}]%
        {Boukes}
\bibfield{author}{\bibinfo{person}{Mark Boukes}.}
  \bibinfo{year}{2019}\natexlab{}.
\newblock \showarticletitle{Social network sites and acquiring current affairs
  knowledge: The impact of Twitter and Facebook usage on learning about the
  news}.
\newblock \bibinfo{journal}{\emph{Journal of Information Technology \&
  Politics}} \bibinfo{volume}{16}, \bibinfo{number}{1} (\bibinfo{year}{2019}),
  \bibinfo{pages}{36--51}.
\newblock
\urldef\tempurl%
\url{https://doi.org/10.1080/19331681.2019.1572568}
\showDOI{\tempurl}


\bibitem[\protect\citeauthoryear{Breiman}{Breiman}{2001}]%
        {Breiman}
\bibfield{author}{\bibinfo{person}{Leo Breiman}.}
  \bibinfo{year}{2001}\natexlab{}.
\newblock \showarticletitle{Random Forests}.
\newblock \bibinfo{journal}{\emph{Machine Learning}} \bibinfo{volume}{45},
  \bibinfo{number}{1} (\bibinfo{date}{Oct} \bibinfo{year}{2001}),
  \bibinfo{pages}{5–32}.
\newblock
\showISSN{1573-0565}
\urldef\tempurl%
\url{https://doi.org/10.1023/A:1010933404324}
\showDOI{\tempurl}


\bibitem[\protect\citeauthoryear{Capistrano, Suarez, and Naval}{Capistrano
  et~al\mbox{.}}{2019}]%
        {Capistrano}
\bibfield{author}{\bibinfo{person}{Jose Lorenzo~C. Capistrano},
  \bibinfo{person}{Jessie James~P. Suarez}, {and} \bibinfo{person}{Prospero~C.
  Naval}.} \bibinfo{year}{2019}\natexlab{}.
\newblock \showarticletitle{SALSA: Detection of Cybertrolls Using Sentiment,
  Aggression, Lexical and Syntactic Analysis of Tweets}. In
  \bibinfo{booktitle}{\emph{Proceedings of the 9th International Conference on
  Web Intelligence, Mining and Semantics}} \emph{(\bibinfo{series}{WIMS2019})}.
  \bibinfo{publisher}{Association for Computing Machinery},
  \bibinfo{address}{New York, NY, USA}, Article \bibinfo{articleno}{10},
  \bibinfo{numpages}{6}~pages.
\newblock
\showISBNx{9781450361903}
\urldef\tempurl%
\url{https://doi.org/10.1145/3326467.3326471}
\showDOI{\tempurl}


\bibitem[\protect\citeauthoryear{Chakraborti and Dey}{Chakraborti and
  Dey}{2014}]%
        {Chakraborti}
\bibfield{author}{\bibinfo{person}{S. Chakraborti} {and} \bibinfo{person}{S.
  Dey}.} \bibinfo{year}{2014}\natexlab{}.
\newblock \showarticletitle{Multi-document Text Summarization for Competitor
  Intelligence: A Methodology}. In \bibinfo{booktitle}{\emph{2014 2nd
  International Symposium on Computational and Business Intelligence}}.
  \bibinfo{pages}{97–100}.
\newblock
\urldef\tempurl%
\url{https://doi.org/10.1109/ISCBI.2014.28}
\showDOI{\tempurl}


\bibitem[\protect\citeauthoryear{Chappell}{Chappell}{2021}]%
        {Chappell}
\bibfield{author}{\bibinfo{person}{Bill Chappell}.}
  \bibinfo{year}{2021}\natexlab{}.
\newblock \bibinfo{title}{{Instagram Bars Robert F. Kennedy Jr. For Spreading
  Vaccine Misinformation}}.
\newblock
\newblock
\newblock
\shownote{\url{https://www.npr.org/sections/coronavirus-live-updates/2021/02/11/966902737/instagram-bars-robert-f-kennedy-jr-for-spreading-vaccine-misinformation}.}


\bibitem[\protect\citeauthoryear{Chen, Lerman, and Ferrara}{Chen
  et~al\mbox{.}}{2020}]%
        {Chen}
\bibfield{author}{\bibinfo{person}{Emily Chen}, \bibinfo{person}{Kristina
  Lerman}, {and} \bibinfo{person}{Emilio Ferrara}.}
  \bibinfo{year}{2020}\natexlab{}.
\newblock \showarticletitle{Tracking Social Media Discourse About the COVID-19
  Pandemic: Development of a Public Coronavirus Twitter Data Set}.
\newblock \bibinfo{journal}{\emph{JMIR Public Health Surveill}}
  \bibinfo{volume}{6}, \bibinfo{number}{2} (\bibinfo{date}{29 May}
  \bibinfo{year}{2020}), \bibinfo{pages}{e19273}.
\newblock
\showISSN{2369-2960}
\urldef\tempurl%
\url{https://doi.org/10.2196/19273}
\showDOI{\tempurl}


\bibitem[\protect\citeauthoryear{Clayton, Blair, Busam, Forstner, Glance,
  Green, Kawata, Kovvuri, Martin, Morgan, et~al\mbox{.}}{Clayton
  et~al\mbox{.}}{2019}]%
        {Clayton}
\bibfield{author}{\bibinfo{person}{Katherine Clayton}, \bibinfo{person}{Spencer
  Blair}, \bibinfo{person}{Jonathan~A Busam}, \bibinfo{person}{Samuel
  Forstner}, \bibinfo{person}{John Glance}, \bibinfo{person}{Guy Green},
  \bibinfo{person}{Anna Kawata}, \bibinfo{person}{Akhila Kovvuri},
  \bibinfo{person}{Jonathan Martin}, \bibinfo{person}{Evan Morgan},
  {et~al\mbox{.}}} \bibinfo{year}{2019}\natexlab{}.
\newblock \showarticletitle{Real solutions for fake news? Measuring the
  effectiveness of general warnings and fact-check tags in reducing belief in
  false stories on social media}.
\newblock \bibinfo{journal}{\emph{Political Behavior}} (\bibinfo{year}{2019}),
  \bibinfo{pages}{1--23}.
\newblock


\bibitem[\protect\citeauthoryear{Dernoncourt, Ghassemi, and Chang}{Dernoncourt
  et~al\mbox{.}}{2018}]%
        {Dernoncourt}
\bibfield{author}{\bibinfo{person}{Franck Dernoncourt},
  \bibinfo{person}{Mohammad Ghassemi}, {and} \bibinfo{person}{Walter Chang}.}
  \bibinfo{year}{2018}\natexlab{}.
\newblock \showarticletitle{A repository of corpora for summarization}. In
  \bibinfo{booktitle}{\emph{Proceedings of the Eleventh International
  Conference on Language Resources and Evaluation (LREC 2018)}}.
\newblock


\bibitem[\protect\citeauthoryear{Desoky, Younis, and El-Sayed}{Desoky
  et~al\mbox{.}}{2008}]%
        {Desoky}
\bibfield{author}{\bibinfo{person}{A. Desoky}, \bibinfo{person}{M. Younis},
  {and} \bibinfo{person}{H. El-Sayed}.} \bibinfo{year}{2008}\natexlab{}.
\newblock \showarticletitle{Auto-Summarization-Based Steganography}. In
  \bibinfo{booktitle}{\emph{2008 International Conference on Innovations in
  Information Technology}}. \bibinfo{pages}{608–612}.
\newblock
\urldef\tempurl%
\url{https://doi.org/10.1109/INNOVATIONS.2008.4781634}
\showDOI{\tempurl}


\bibitem[\protect\citeauthoryear{El-Kassas, Salama, Rafea, and
  Mohamed}{El-Kassas et~al\mbox{.}}{2021}]%
        {El-Kassas}
\bibfield{author}{\bibinfo{person}{Wafaa~S. El-Kassas},
  \bibinfo{person}{Cherif~R. Salama}, \bibinfo{person}{Ahmed~A. Rafea}, {and}
  \bibinfo{person}{Hoda~K. Mohamed}.} \bibinfo{year}{2021}\natexlab{}.
\newblock \showarticletitle{Automatic text summarization: A comprehensive
  survey}.
\newblock \bibinfo{journal}{\emph{Expert Systems with Applications}}
  \bibinfo{volume}{165} (\bibinfo{year}{2021}), \bibinfo{pages}{113679}.
\newblock
\showISSN{0957-4174}
\urldef\tempurl%
\url{https://doi.org/10.1016/j.eswa.2020.113679}
\showDOI{\tempurl}


\bibitem[\protect\citeauthoryear{Ferreira, Freitas, Cabral, Lins, Lima,
  França, Simske, and Favaro}{Ferreira et~al\mbox{.}}{2014}]%
        {Ferreira}
\bibfield{author}{\bibinfo{person}{R. Ferreira}, \bibinfo{person}{F. Freitas},
  \bibinfo{person}{L.~d~S. Cabral}, \bibinfo{person}{R.~D. Lins},
  \bibinfo{person}{R. Lima}, \bibinfo{person}{G. França},
  \bibinfo{person}{S.~J. Simske}, {and} \bibinfo{person}{L. Favaro}.}
  \bibinfo{year}{2014}\natexlab{}.
\newblock \showarticletitle{A Context Based Text Summarization System}. In
  \bibinfo{booktitle}{\emph{2014 11th IAPR International Workshop on Document
  Analysis Systems}}. \bibinfo{pages}{66–70}.
\newblock
\urldef\tempurl%
\url{https://doi.org/10.1109/DAS.2014.19}
\showDOI{\tempurl}


\bibitem[\protect\citeauthoryear{for Disease~Control and Prevention}{for
  Disease~Control and Prevention}{2021}]%
        {CDC}
\bibfield{author}{\bibinfo{person}{Centers for Disease~Control} {and}
  \bibinfo{person}{Prevention}.} \bibinfo{year}{2021}\natexlab{}.
\newblock \bibinfo{title}{{COVID-19 Vaccines and Allergic Reactions}}.
\newblock
\newblock
\newblock
\shownote{\url{https://www.cdc.gov/coronavirus/2019-ncov/vaccines/safety/allergic-reaction.html}.}


\bibitem[\protect\citeauthoryear{Fornacciari, Mordonini, Poggi, Sani, and
  Tomaiuolo}{Fornacciari et~al\mbox{.}}{2018}]%
        {Fornacciari}
\bibfield{author}{\bibinfo{person}{Paolo Fornacciari}, \bibinfo{person}{Monica
  Mordonini}, \bibinfo{person}{Agostino Poggi}, \bibinfo{person}{Laura Sani},
  {and} \bibinfo{person}{Michele Tomaiuolo}.} \bibinfo{year}{2018}\natexlab{}.
\newblock \showarticletitle{A holistic system for troll detection on Twitter}.
\newblock \bibinfo{journal}{\emph{Computers in Human Behavior}}
  \bibinfo{volume}{89} (\bibinfo{year}{2018}), \bibinfo{pages}{258 -- 268}.
\newblock
\showISSN{0747-5632}
\urldef\tempurl%
\url{https://doi.org/10.1016/j.chb.2018.08.008}
\showDOI{\tempurl}


\bibitem[\protect\citeauthoryear{Freelon, Marwick, and Kreiss}{Freelon
  et~al\mbox{.}}{2020}]%
        {Freelon}
\bibfield{author}{\bibinfo{person}{Deen Freelon}, \bibinfo{person}{Alice
  Marwick}, {and} \bibinfo{person}{Daniel Kreiss}.}
  \bibinfo{year}{2020}\natexlab{}.
\newblock \showarticletitle{False equivalencies: Online activism from left to
  right}.
\newblock \bibinfo{journal}{\emph{Science}} \bibinfo{volume}{369},
  \bibinfo{number}{6508} (\bibinfo{year}{2020}), \bibinfo{pages}{1197--1201}.
\newblock
\showISSN{0036-8075}
\urldef\tempurl%
\url{https://doi.org/10.1126/science.abb2428}
\showDOI{\tempurl}


\bibitem[\protect\citeauthoryear{Frenkel, Abi-Habib, and Barnes}{Frenkel
  et~al\mbox{.}}{2021}]%
        {Frenkel}
\bibfield{author}{\bibinfo{person}{Sheera Frenkel}, \bibinfo{person}{Maria
  Abi-Habib}, {and} \bibinfo{person}{Julian~E. Barnes}.}
  \bibinfo{year}{2021}\natexlab{}.
\newblock \bibinfo{title}{{Russian Campaign Promotes Homegrown Vaccine and
  Undercuts Rivals}}.
\newblock
\newblock
\newblock
\shownote{\url{https://www.nytimes.com/2021/02/05/technology/russia-covid-vaccine-disinformation.html}.}


\bibitem[\protect\citeauthoryear{Gambhir and Gupta}{Gambhir and Gupta}{2017}]%
        {Mahak}
\bibfield{author}{\bibinfo{person}{Mahak Gambhir} {and} \bibinfo{person}{Vishal
  Gupta}.} \bibinfo{year}{2017}\natexlab{}.
\newblock \showarticletitle{Recent automatic text summarization techniques: a
  survey}.
\newblock \bibinfo{journal}{\emph{Artificial Intelligence Review}}
  \bibinfo{volume}{47}, \bibinfo{number}{1} (\bibinfo{year}{2017}),
  \bibinfo{pages}{1--66}.
\newblock
\showISBNx{1573-7462}
\urldef\tempurl%
\url{https://doi.org/10.1007/s10462-016-9475-9}
\showDOI{\tempurl}


\bibitem[\protect\citeauthoryear{Geeng, Yee, and Roesner}{Geeng
  et~al\mbox{.}}{2020}]%
        {Geeng}
\bibfield{author}{\bibinfo{person}{Christine Geeng}, \bibinfo{person}{Savanna
  Yee}, {and} \bibinfo{person}{Franziska Roesner}.}
  \bibinfo{year}{2020}\natexlab{}.
\newblock \showarticletitle{Fake News on Facebook and Twitter: Investigating
  How People (Don't) Investigate}. In \bibinfo{booktitle}{\emph{Proceedings of
  the 2020 CHI Conference on Human Factors in Computing Systems}}
  \emph{(\bibinfo{series}{CHI '20})}. \bibinfo{publisher}{Association for
  Computing Machinery}, \bibinfo{address}{New York, NY, USA},
  \bibinfo{pages}{1–14}.
\newblock
\showISBNx{9781450367080}
\urldef\tempurl%
\url{https://doi.org/10.1145/3313831.3376784}
\showDOI{\tempurl}


\bibitem[\protect\citeauthoryear{Graham, Haidt, and Nosek}{Graham
  et~al\mbox{.}}{2009}]%
        {Graham}
\bibfield{author}{\bibinfo{person}{Jesse Graham}, \bibinfo{person}{Jonathan
  Haidt}, {and} \bibinfo{person}{Brian~A Nosek}.}
  \bibinfo{year}{2009}\natexlab{}.
\newblock \showarticletitle{Liberals and conservatives rely on different sets
  of moral foundations.}
\newblock \bibinfo{journal}{\emph{Journal of personality and social
  psychology}} \bibinfo{volume}{96}, \bibinfo{number}{5}
  (\bibinfo{year}{2009}), \bibinfo{pages}{1029}.
\newblock


\bibitem[\protect\citeauthoryear{Hanyok}{Hanyok}{2005}]%
        {nsa}
\bibfield{author}{\bibinfo{person}{Robert Hanyok}.}
  \bibinfo{year}{2005}\natexlab{}.
\newblock \bibinfo{title}{Skunks, Bogies, Silent Hounds, and the Flying Fish:
  The Gulf of Tonkin Mystery, 2-4 August 1964}.
\newblock
\newblock
\urldef\tempurl%
\url{https://www.nsa.gov/Portals/70/documents/news-features/declassified-documents/gulf-of-tonkin/articles/release-1/rel1_skunks_bogies.pdf}
\showURL{%
\tempurl}


\bibitem[\protect\citeauthoryear{{Higashinaka}, {Kawamae}, {Sadamitsu},
  {Minami}, {Meguro}, {Dohsaka}, and {Inagaki}}{{Higashinaka}
  et~al\mbox{.}}{2011}]%
        {Higashinaka}
\bibfield{author}{\bibinfo{person}{R. {Higashinaka}}, \bibinfo{person}{N.
  {Kawamae}}, \bibinfo{person}{K. {Sadamitsu}}, \bibinfo{person}{Y. {Minami}},
  \bibinfo{person}{T. {Meguro}}, \bibinfo{person}{K. {Dohsaka}}, {and}
  \bibinfo{person}{H. {Inagaki}}.} \bibinfo{year}{2011}\natexlab{}.
\newblock \showarticletitle{Building a conversational model from two-tweets}.
  In \bibinfo{booktitle}{\emph{2011 IEEE Workshop on Automatic Speech
  Recognition Understanding}}. \bibinfo{pages}{330--335}.
\newblock
\urldef\tempurl%
\url{https://doi.org/10.1109/ASRU.2011.6163953}
\showDOI{\tempurl}


\bibitem[\protect\citeauthoryear{Himelboim, Sweetser, Tinkham, Cameron, Danelo,
  and West}{Himelboim et~al\mbox{.}}{2016}]%
        {Himelboim}
\bibfield{author}{\bibinfo{person}{Itai Himelboim}, \bibinfo{person}{Kaye~D
  Sweetser}, \bibinfo{person}{Spencer~F Tinkham}, \bibinfo{person}{Kristen
  Cameron}, \bibinfo{person}{Matthew Danelo}, {and} \bibinfo{person}{Kate
  West}.} \bibinfo{year}{2016}\natexlab{}.
\newblock \showarticletitle{Valence-based homophily on Twitter: Network
  Analysis of Emotions and Political Talk in the 2012 Presidential Election}.
\newblock \bibinfo{journal}{\emph{New Media \& Society}} \bibinfo{volume}{18},
  \bibinfo{number}{7} (\bibinfo{year}{2016}), \bibinfo{pages}{1382--1400}.
\newblock
\urldef\tempurl%
\url{https://doi.org/10.1177/1461444814555096}
\showDOI{\tempurl}


\bibitem[\protect\citeauthoryear{Holmes}{Holmes}{2007}]%
        {Holmes}
\bibfield{author}{\bibinfo{person}{David~G. Holmes}.}
  \bibinfo{year}{2007}\natexlab{}.
\newblock \showarticletitle{Affirmative Reaction: Kennedy, Nixon, King, and the
  Evolution of Color-Blind Rhetoric}.
\newblock \bibinfo{journal}{\emph{Rhetoric Review}} \bibinfo{volume}{26},
  \bibinfo{number}{1} (\bibinfo{year}{2007}), \bibinfo{pages}{25--41}.
\newblock
\showISSN{07350198, 15327981}
\urldef\tempurl%
\url{http://www.jstor.org/stable/20176758}
\showURL{%
\tempurl}


\bibitem[\protect\citeauthoryear{Jachim, Sharevski, and Pieroni}{Jachim
  et~al\mbox{.}}{2021}]%
        {iwspa2021}
\bibfield{author}{\bibinfo{person}{Peter Jachim}, \bibinfo{person}{Filipo
  Sharevski}, {and} \bibinfo{person}{Emma Pieroni}.}
  \bibinfo{year}{2021}\natexlab{}.
\newblock \showarticletitle{TrollHunter2020: Real-time Detection of Trolling
  Narratives on Twitter During the 2020 US Elections}. In
  \bibinfo{booktitle}{\emph{International Workshop on Security and Privacy
  Analytics 2021}} \emph{(\bibinfo{series}{IWSPA '21})}.
  \bibinfo{publisher}{Association for Computing Machinery},
  \bibinfo{address}{New York, NY, USA}, \bibinfo{pages}{1--11}.
\newblock
\urldef\tempurl%
\url{https://doi.org/10.1145/3445970.3451158}
\showDOI{\tempurl}


\bibitem[\protect\citeauthoryear{Jachim, Sharevski, and Treebridge}{Jachim
  et~al\mbox{.}}{2020}]%
        {nspw2020}
\bibfield{author}{\bibinfo{person}{Peter Jachim}, \bibinfo{person}{Filipo
  Sharevski}, {and} \bibinfo{person}{Paige Treebridge}.}
  \bibinfo{year}{2020}\natexlab{}.
\newblock \showarticletitle{TrollHunter [Evader]: Automated Detection [Evasion]
  of Twitter Trolls During the COVID-19 Pandemic}. In
  \bibinfo{booktitle}{\emph{New Security Paradigms Workshop 2020}}
  \emph{(\bibinfo{series}{NSPW '20})}. \bibinfo{publisher}{Association for
  Computing Machinery}, \bibinfo{address}{New York, NY, USA},
  \bibinfo{pages}{59–75}.
\newblock
\showISBNx{9781450389952}
\urldef\tempurl%
\url{https://doi.org/10.1145/3442167.3442169}
\showDOI{\tempurl}


\bibitem[\protect\citeauthoryear{Jones}{Jones}{2007}]%
        {Jones}
\bibfield{author}{\bibinfo{person}{Karen~Sp{\"a}rck Jones}.}
  \bibinfo{year}{2007}\natexlab{}.
\newblock \showarticletitle{Automatic summarising: The state of the art}.
\newblock \bibinfo{journal}{\emph{Information Processing \& Management}}
  \bibinfo{volume}{43}, \bibinfo{number}{6} (\bibinfo{year}{2007}),
  \bibinfo{pages}{1449--1481}.
\newblock
\showISSN{0306-4573}
\urldef\tempurl%
\url{https://doi.org/10.1016/j.ipm.2007.03.009}
\showDOI{\tempurl}
\newblock
\shownote{Text Summarization.}


\bibitem[\protect\citeauthoryear{Keskar, McCann, Varshney, Xiong, and
  Socher}{Keskar et~al\mbox{.}}{2019}]%
        {Keskar}
\bibfield{author}{\bibinfo{person}{Nitish~Shirish Keskar},
  \bibinfo{person}{Bryan McCann}, \bibinfo{person}{Lav~R. Varshney},
  \bibinfo{person}{Caiming Xiong}, {and} \bibinfo{person}{Richard Socher}.}
  \bibinfo{year}{2019}\natexlab{}.
\newblock \bibinfo{title}{CTRL: A Conditional Transformer Language Model for
  Controllable Generation}.
\newblock
\newblock
\showeprint[arxiv]{cs.CL/1909.05858}


\bibitem[\protect\citeauthoryear{Korobov and Lopuhin}{Korobov and Lopuhin}{[n.
  d.]}]%
        {eli5}
\bibfield{author}{\bibinfo{person}{Mikhail Korobov} {and}
  \bibinfo{person}{Konstantin Lopuhin}.} \bibinfo{year}{[n. d.]}\natexlab{}.
\newblock \bibinfo{booktitle}{\emph{ELI5}}.
\newblock \bibinfo{publisher}{Open Source}.
\newblock
\urldef\tempurl%
\url{https://eli5.readthedocs.io/}
\showURL{%
\tempurl}


\bibitem[\protect\citeauthoryear{Marcos}{Marcos}{2021}]%
        {COVIDSumm}
\bibfield{author}{\bibinfo{person}{Cristina Marcos}.}
  \bibinfo{year}{2021}\natexlab{}.
\newblock \bibinfo{title}{No Republicans back \$1.9T COVID-19 Relief Bill}.
\newblock
\newblock
\urldef\tempurl%
\url{https://thehill.com/homenews/house/542581-no-republicans-back-19t-covid-19-relief-bill}
\showURL{%
\tempurl}


\bibitem[\protect\citeauthoryear{McGee}{McGee}{2021}]%
        {McGee}
\bibfield{author}{\bibinfo{person}{Meredith McGee}.}
  \bibinfo{year}{2021}\natexlab{}.
\newblock \bibinfo{title}{{COVID-19 Vaccines and Allergic Reactions}}.
\newblock
\newblock
\newblock
\shownote{\url{https://thehill.com/opinion/campaign/540541-why-congress-should-provide-sustained-funding-in-elections?rl=1}.}


\bibitem[\protect\citeauthoryear{McGlone}{McGlone}{2005}]%
        {McGlone}
\bibfield{author}{\bibinfo{person}{Matthew~S. McGlone}.}
  \bibinfo{year}{2005}\natexlab{}.
\newblock \showarticletitle{Contextomy: the art of quoting out of context}.
\newblock \bibinfo{journal}{\emph{Media, Culture \& Society}}
  \bibinfo{volume}{27}, \bibinfo{number}{4} (\bibinfo{year}{2005}),
  \bibinfo{pages}{511--522}.
\newblock
\urldef\tempurl%
\url{https://doi.org/10.1177/0163443705053974}
\showDOI{\tempurl}


\bibitem[\protect\citeauthoryear{Morris, Lifland, Yoo, Grigsby, Jin, and
  Qi}{Morris et~al\mbox{.}}{2020}]%
        {Morris}
\bibfield{author}{\bibinfo{person}{John~X. Morris}, \bibinfo{person}{Eli
  Lifland}, \bibinfo{person}{Jin~Yong Yoo}, \bibinfo{person}{Jake Grigsby},
  \bibinfo{person}{Di Jin}, {and} \bibinfo{person}{Yanjun Qi}.}
  \bibinfo{year}{2020}\natexlab{}.
\newblock \bibinfo{title}{TextAttack: A Framework for Adversarial Attacks, Data
  Augmentation, and Adversarial Training in NLP}.
\newblock
\newblock
\showeprint[arxiv]{cs.CL/2005.05909}


\bibitem[\protect\citeauthoryear{Nyhan and Reifler}{Nyhan and Reifler}{2010}]%
        {Nyhan}
\bibfield{author}{\bibinfo{person}{Brendan Nyhan} {and} \bibinfo{person}{Jason
  Reifler}.} \bibinfo{year}{2010}\natexlab{}.
\newblock \showarticletitle{When corrections fail: The persistence of political
  misperceptions}.
\newblock \bibinfo{journal}{\emph{Political Behavior}} \bibinfo{volume}{32},
  \bibinfo{number}{2} (\bibinfo{year}{2010}), \bibinfo{pages}{303--330}.
\newblock


\bibitem[\protect\citeauthoryear{Ortutay}{Ortutay}{2021}]%
        {Ortutay2021}
\bibfield{author}{\bibinfo{person}{Barbara Ortutay}.}
  \bibinfo{year}{2021}\natexlab{}.
\newblock \showarticletitle{Twitter launches crowd-sourced fact-checking
  project}.
\newblock \bibinfo{journal}{\emph{Associated Press - AP News}}
  (\bibinfo{year}{2021}).
\newblock
\urldef\tempurl%
\url{https://apnews.com/article/twitter-launch-crowd-sourced-fact-check-589809d4c9a7eceda1ea8293b0a14af2}
\showURL{%
\tempurl}


\bibitem[\protect\citeauthoryear{Pancer and Poole}{Pancer and Poole}{2016}]%
        {Pancer}
\bibfield{author}{\bibinfo{person}{Ethan Pancer} {and} \bibinfo{person}{Maxwell
  Poole}.} \bibinfo{year}{2016}\natexlab{}.
\newblock \showarticletitle{The popularity and virality of political social
  media: hashtags, mentions, and links predict likes and retweets of 2016 U.S.
  presidential nominees' tweets}.
\newblock \bibinfo{journal}{\emph{Social Influence}} \bibinfo{volume}{11},
  \bibinfo{number}{4} (\bibinfo{year}{2016}), \bibinfo{pages}{259--270}.
\newblock
\urldef\tempurl%
\url{https://doi.org/10.1080/15534510.2016.1265582}
\showDOI{\tempurl}
\showeprint{https://doi.org/10.1080/15534510.2016.1265582}


\bibitem[\protect\citeauthoryear{Pedregosa, Varoquaux, Gramfort, Michel,
  Thirion, Grisel, Blondel, Prettenhofer, Weiss, Dubourg, Vanderplas, Passos,
  Cournapeau, Brucher, Perrot, and Duchesnay}{Pedregosa et~al\mbox{.}}{2011}]%
        {sklearn}
\bibfield{author}{\bibinfo{person}{F. Pedregosa}, \bibinfo{person}{G.
  Varoquaux}, \bibinfo{person}{A. Gramfort}, \bibinfo{person}{V. Michel},
  \bibinfo{person}{B. Thirion}, \bibinfo{person}{O. Grisel},
  \bibinfo{person}{M. Blondel}, \bibinfo{person}{P. Prettenhofer},
  \bibinfo{person}{R. Weiss}, \bibinfo{person}{V. Dubourg}, \bibinfo{person}{J.
  Vanderplas}, \bibinfo{person}{A. Passos}, \bibinfo{person}{D. Cournapeau},
  \bibinfo{person}{M. Brucher}, \bibinfo{person}{M. Perrot}, {and}
  \bibinfo{person}{E. Duchesnay}.} \bibinfo{year}{2011}\natexlab{}.
\newblock \showarticletitle{Scikit-learn: Machine Learning in {P}ython}.
\newblock \bibinfo{journal}{\emph{Journal of Machine Learning Research}}
  \bibinfo{volume}{12} (\bibinfo{year}{2011}), \bibinfo{pages}{2825--2830}.
\newblock


\bibitem[\protect\citeauthoryear{Pennebaker, Francis, and Booth}{Pennebaker
  et~al\mbox{.}}{2001}]%
        {Pennebaker}
\bibfield{author}{\bibinfo{person}{James~W Pennebaker},
  \bibinfo{person}{Martha~E Francis}, {and} \bibinfo{person}{Roger~J Booth}.}
  \bibinfo{year}{2001}\natexlab{}.
\newblock \showarticletitle{Linguistic inquiry and word count: LIWC 2001}.
\newblock \bibinfo{journal}{\emph{Mahway: Lawrence Erlbaum Associates}}
  \bibinfo{volume}{71}, \bibinfo{number}{2001} (\bibinfo{year}{2001}),
  \bibinfo{pages}{2001}.
\newblock


\bibitem[\protect\citeauthoryear{Pennycook, Bear, Collins, and Rand}{Pennycook
  et~al\mbox{.}}{2020}]%
        {Pennycook}
\bibfield{author}{\bibinfo{person}{Gordon Pennycook}, \bibinfo{person}{Adam
  Bear}, \bibinfo{person}{Evan~T Collins}, {and} \bibinfo{person}{David~G
  Rand}.} \bibinfo{year}{2020}\natexlab{}.
\newblock \showarticletitle{The implied truth effect: Attaching warnings to a
  subset of fake news headlines increases perceived accuracy of headlines
  without warnings}.
\newblock \bibinfo{journal}{\emph{Management Science}} (\bibinfo{year}{2020}).
\newblock


\bibitem[\protect\citeauthoryear{Pennycook, Cannon, and Rand}{Pennycook
  et~al\mbox{.}}{2018}]%
        {Pennycook1}
\bibfield{author}{\bibinfo{person}{Gordon Pennycook}, \bibinfo{person}{Tyrone~D
  Cannon}, {and} \bibinfo{person}{David~G Rand}.}
  \bibinfo{year}{2018}\natexlab{}.
\newblock \showarticletitle{Prior exposure increases perceived accuracy of fake
  news.}
\newblock \bibinfo{journal}{\emph{Journal of experimental psychology: general}}
  \bibinfo{volume}{147}, \bibinfo{number}{12} (\bibinfo{year}{2018}),
  \bibinfo{pages}{1865}.
\newblock


\bibitem[\protect\citeauthoryear{Pieroni, Jachim, Jachim, and
  Sharevski}{Pieroni et~al\mbox{.}}{2021}]%
        {chi2021}
\bibfield{author}{\bibinfo{person}{Emma Pieroni}, \bibinfo{person}{Peter
  Jachim}, \bibinfo{person}{Nathaniel Jachim}, {and} \bibinfo{person}{Filipo
  Sharevski}.} \bibinfo{year}{2021}\natexlab{}.
\newblock \showarticletitle{Parlermonium: A Data-Driven UX Design Evaluation of
  the Parler Platform}. In \bibinfo{booktitle}{\emph{CHI 2021 - Technologies
  for Support of Critical Thinking in the Age of Misinformation}}
  \emph{(\bibinfo{series}{CHI Workshops '21})}. \bibinfo{publisher}{Association
  for Computing Machinery}, \bibinfo{address}{New York, NY, USA},
  \bibinfo{pages}{1--10}.
\newblock


\bibitem[\protect\citeauthoryear{Radford, Wu, Child, Luan, Amodei, and
  Sutskever}{Radford et~al\mbox{.}}{[n. d.]}]%
        {Radford}
\bibfield{author}{\bibinfo{person}{Alec Radford}, \bibinfo{person}{Jeffrey Wu},
  \bibinfo{person}{Rewon Child}, \bibinfo{person}{David Luan},
  \bibinfo{person}{Dario Amodei}, {and} \bibinfo{person}{Ilya Sutskever}.}
  \bibinfo{year}{[n. d.]}\natexlab{}.
\newblock \showarticletitle{Language models are unsupervised multitask
  learners}.
\newblock  (\bibinfo{year}{[n. d.]}).
\newblock


\bibitem[\protect\citeauthoryear{Ribeiro, Singh, and Guestrin}{Ribeiro
  et~al\mbox{.}}{2016}]%
        {ribeiro}
\bibfield{author}{\bibinfo{person}{Marco~Tulio Ribeiro},
  \bibinfo{person}{Sameer Singh}, {and} \bibinfo{person}{Carlos Guestrin}.}
  \bibinfo{year}{2016}\natexlab{}.
\newblock \showarticletitle{“Why Should I Trust You?”: Explaining the
  Predictions of Any Classifier}.
\newblock \bibinfo{journal}{\emph{arXiv:1602.04938 [cs, stat]}}
  (\bibinfo{date}{Aug} \bibinfo{year}{2016}).
\newblock
\urldef\tempurl%
\url{http://arxiv.org/abs/1602.04938}
\showURL{%
\tempurl}
\newblock
\shownote{arXiv: 1602.04938.}


\bibitem[\protect\citeauthoryear{Richardson}{Richardson}{2004}]%
        {bs4}
\bibfield{author}{\bibinfo{person}{Leonard Richardson}.}
  \bibinfo{year}{2004}\natexlab{}.
\newblock \bibinfo{booktitle}{\emph{Beautiful Soup}}.
\newblock
\urldef\tempurl%
\url{https://www.crummy.com/software/BeautifulSoup/bs4/doc/}
\showURL{%
\tempurl}


\bibitem[\protect\citeauthoryear{Safety}{Safety}{2021}]%
        {TwitterUpdate}
\bibfield{author}{\bibinfo{person}{Twitter Safety}.}
  \bibinfo{year}{2021}\natexlab{}.
\newblock \showarticletitle{Updates to our work on COVID-19 vaccine
  misinformation}.
\newblock  (\bibinfo{date}{March} \bibinfo{year}{2021}).
\newblock
\urldef\tempurl%
\url{https://blog.twitter.com/en_us/topics/company/2021/updates-to-our-work-on-covid-19-vaccine-misinformation.html}
\showURL{%
\tempurl}


\bibitem[\protect\citeauthoryear{Sharevski, Alsaadi, Jachim, and
  Pieroni}{Sharevski et~al\mbox{.}}{2021a}]%
        {soups2021}
\bibfield{author}{\bibinfo{person}{Filipo Sharevski}, \bibinfo{person}{Ranniem
  Alsaadi}, \bibinfo{person}{Peter Jachim}, {and} \bibinfo{person}{Emma
  Pieroni}.} \bibinfo{year}{2021}\natexlab{a}.
\newblock \showarticletitle{COVID-19 Misinformation Warning Labels: Twitter's
  Soft Moderation Effects on Belief Echoes about the COVID-19 Vaccine}. In
  \bibinfo{booktitle}{\emph{Seventeenth Symposium on Usable Privacy and
  Security ({SOUPS} 2021)}}. \bibinfo{publisher}{{USENIX} Association}.
\newblock
\urldef\tempurl%
\url{https://www.usenix.org/conference/soups2021}
\showURL{%
\tempurl}


\bibitem[\protect\citeauthoryear{Sharevski, Jachim, and Pieroni}{Sharevski
  et~al\mbox{.}}{2021b}]%
        {acsac2020}
\bibfield{author}{\bibinfo{person}{Filipo Sharevski}, \bibinfo{person}{Peter
  Jachim}, {and} \bibinfo{person}{Emma Pieroni}.}
  \bibinfo{year}{2021}\natexlab{b}.
\newblock \showarticletitle{WikipediaBot: Machine Learning Assisted Adversarial
  Manipulation of Wikipedia Articles}. In \bibinfo{booktitle}{\emph{DYnamic and
  Novel Advances in Machine Learning and Intelligent Cyber Security, (DYNAMICS)
  Workshop, Annual Computer Security Applications Conference (ACSAC) 2020}}
  \emph{(\bibinfo{series}{ACSAC '20})}. \bibinfo{publisher}{Association for
  Computing Machinery}, \bibinfo{address}{New York, NY, USA},
  \bibinfo{pages}{1--8}.
\newblock


\bibitem[\protect\citeauthoryear{{Sharevski}, {Jachim}, {Treebridge}, {Li}, and
  {Babin}}{{Sharevski} et~al\mbox{.}}{2020}]%
        {eurosp2020}
\bibfield{author}{\bibinfo{person}{F. {Sharevski}}, \bibinfo{person}{P.
  {Jachim}}, \bibinfo{person}{P. {Treebridge}}, \bibinfo{person}{A. {Li}},
  {and} \bibinfo{person}{A. {Babin}}.} \bibinfo{year}{2020}\natexlab{}.
\newblock \showarticletitle{My Boss is Really Cool: Malware-Induced
  Misperception in Workplace Communication Through Covert Linguistic
  Manipulation of Emails}. In \bibinfo{booktitle}{\emph{2020 IEEE European
  Symposium on Security and Privacy Workshops (EuroS\&PW)}}.
  \bibinfo{pages}{463--470}.
\newblock
\urldef\tempurl%
\url{https://doi.org/10.1109/EuroSPW51379.2020.00068}
\showDOI{\tempurl}


\bibitem[\protect\citeauthoryear{Sharevski, Jachim, Treebridge, Li, Babin, and
  Adadevoh}{Sharevski et~al\mbox{.}}{2021c}]%
        {malexa2021}
\bibfield{author}{\bibinfo{person}{Filipo Sharevski}, \bibinfo{person}{Peter
  Jachim}, \bibinfo{person}{Paige Treebridge}, \bibinfo{person}{Audrey Li},
  \bibinfo{person}{Adam Babin}, {and} \bibinfo{person}{Christopher Adadevoh}.}
  \bibinfo{year}{2021}\natexlab{c}.
\newblock \showarticletitle{Meet Malexa, Alexa's malicious twin:
  Malware-induced misperception through intelligent voice assistants}.
\newblock \bibinfo{journal}{\emph{International Journal of Human-Computer
  Studies}}  \bibinfo{volume}{149} (\bibinfo{year}{2021}),
  \bibinfo{pages}{102604}.
\newblock
\showISSN{1071-5819}
\urldef\tempurl%
\url{https://doi.org/10.1016/j.ijhcs.2021.102604}
\showDOI{\tempurl}


\bibitem[\protect\citeauthoryear{Sharevski, Slowinski, Jachim, and
  Pieroni}{Sharevski et~al\mbox{.}}{2021d}]%
        {dimwa2021}
\bibfield{author}{\bibinfo{person}{Filipo Sharevski}, \bibinfo{person}{Anna
  Slowinski}, \bibinfo{person}{Peter Jachim}, {and} \bibinfo{person}{Emma
  Pieroni}.} \bibinfo{year}{2021}\natexlab{d}.
\newblock \showarticletitle{``Hey Alexa, What do You Know About the COVID-19
  Vaccine?'' - (Mis)perceptions of Mass Immunization Among Voice Assistant
  Users}. In \bibinfo{booktitle}{\emph{Detection of Intrusions and Malware, and
  Vulnerability Assessment}}. \bibinfo{publisher}{Springer International
  Publishing}, \bibinfo{address}{Chicago, IL}, \bibinfo{pages}{1--11}.
\newblock


\bibitem[\protect\citeauthoryear{Shevlane and Dafoe}{Shevlane and
  Dafoe}{2020}]%
        {Shevlane}
\bibfield{author}{\bibinfo{person}{Toby Shevlane} {and} \bibinfo{person}{Allan
  Dafoe}.} \bibinfo{year}{2020}\natexlab{}.
\newblock \showarticletitle{The Offense-Defense Balance of Scientific
  Knowledge: Does Publishing AI Research Reduce Misuse?}. In
  \bibinfo{booktitle}{\emph{Proceedings of the AAAI/ACM Conference on AI,
  Ethics, and Society}} \emph{(\bibinfo{series}{AIES '20})}.
  \bibinfo{publisher}{Association for Computing Machinery},
  \bibinfo{address}{New York, NY, USA}, \bibinfo{pages}{173–179}.
\newblock
\showISBNx{9781450371100}
\urldef\tempurl%
\url{https://doi.org/10.1145/3375627.3375815}
\showDOI{\tempurl}


\bibitem[\protect\citeauthoryear{Sklearn}{Sklearn}{2021}]%
        {RandomForestClassifier}
\bibfield{author}{\bibinfo{person}{Sklearn}.} \bibinfo{year}{2021}\natexlab{}.
\newblock \bibinfo{title}{{sklearn.ensemble.RandomForestClassifier —
  scikit-learn 0.24.1 documentation\_2021}}.
\newblock
\newblock
\urldef\tempurl%
\url{https://scikit-learn.org/stable/modules/generated/sklearn.ensemble.RandomForestClassifier.html}
\showURL{%
\tempurl}


\bibitem[\protect\citeauthoryear{Staff}{Staff}{2020}]%
        {TwitterUpdate2}
\bibfield{author}{\bibinfo{person}{Reuters Staff}.}
  \bibinfo{year}{2020}\natexlab{}.
\newblock \showarticletitle{Fact check: Twitter’s ‘read before you
  retweet’ warnings do not just target conservative articles}.
\newblock  (\bibinfo{date}{October} \bibinfo{year}{2020}).
\newblock
\urldef\tempurl%
\url{https://www.reuters.com/article/uk-factcheck-twitter-warning-conservativ-idUSKBN2781UK}
\showURL{%
\tempurl}


\bibitem[\protect\citeauthoryear{Starbird}{Starbird}{2017}]%
        {Starbird}
\bibfield{author}{\bibinfo{person}{Kate Starbird}.}
  \bibinfo{year}{2017}\natexlab{}.
\newblock \showarticletitle{Examining the alternative media ecosystem through
  the production of alternative narratives of mass shooting events on Twitter.}
\newblock


\bibitem[\protect\citeauthoryear{Sterling, Jost, and Bonneau}{Sterling
  et~al\mbox{.}}{2020}]%
        {Sterling}
\bibfield{author}{\bibinfo{person}{Joanna Sterling}, \bibinfo{person}{John~T
  Jost}, {and} \bibinfo{person}{Richard Bonneau}.}
  \bibinfo{year}{2020}\natexlab{}.
\newblock \showarticletitle{Political psycholinguistics: A comprehensive
  analysis of the language habits of liberal and conservative social media
  users.}
\newblock \bibinfo{journal}{\emph{Journal of personality and social
  psychology}} (\bibinfo{year}{2020}).
\newblock


\bibitem[\protect\citeauthoryear{Stewart, Arif, and Starbird}{Stewart
  et~al\mbox{.}}{2018}]%
        {Stewart}
\bibfield{author}{\bibinfo{person}{Leo~G Stewart}, \bibinfo{person}{Ahmer
  Arif}, {and} \bibinfo{person}{Kate Starbird}.}
  \bibinfo{year}{2018}\natexlab{}.
\newblock \showarticletitle{Examining trolls and polarization with a retweet
  network}. In \bibinfo{booktitle}{\emph{Proc. ACM WSDM, workshop on
  misinformation and misbehavior mining on the web}}.
\newblock


\bibitem[\protect\citeauthoryear{Tillman}{Tillman}{2021}]%
        {BuzzSumm}
\bibfield{author}{\bibinfo{person}{Zoe Tillman}.}
  \bibinfo{year}{2021}\natexlab{}.
\newblock \bibinfo{title}{Trump Has Now Officially Lost All Of His Postelection
  Challenges In The Supreme Court}.
\newblock
\newblock
\urldef\tempurl%
\url{https://www.buzzfeednews.com/article/zoetillman/trump-supreme-court-election-loss}
\showURL{%
\tempurl}


\bibitem[\protect\citeauthoryear{Twitter}{Twitter}{2020}]%
        {Twitter}
\bibfield{author}{\bibinfo{person}{Twitter}.} \bibinfo{year}{2020}\natexlab{}.
\newblock \bibinfo{title}{Information Operations}.
\newblock
\newblock
\urldef\tempurl%
\url{https://transparency.twitter.com/en/reports/information-operations.html}
\showURL{%
\tempurl}


\bibitem[\protect\citeauthoryear{Wihbey, Coleman, Joseph, and Lazer}{Wihbey
  et~al\mbox{.}}{2017}]%
        {Wihbey}
\bibfield{author}{\bibinfo{person}{John Wihbey}, \bibinfo{person}{Thalita~Dias
  Coleman}, \bibinfo{person}{Kenneth Joseph}, {and} \bibinfo{person}{David
  Lazer}.} \bibinfo{year}{2017}\natexlab{}.
\newblock \bibinfo{title}{Exploring the Ideological Nature of Journalists'
  Social Networks on Twitter and Associations with News Story Content}.
\newblock
\newblock
\showeprint[arxiv]{cs.SI/1708.06727}


\bibitem[\protect\citeauthoryear{Wojcik, Hovasapian, Graham, Motyl, and
  Ditto}{Wojcik et~al\mbox{.}}{2015}]%
        {Wojcik}
\bibfield{author}{\bibinfo{person}{Sean~P. Wojcik}, \bibinfo{person}{Arpine
  Hovasapian}, \bibinfo{person}{Jesse Graham}, \bibinfo{person}{Matt Motyl},
  {and} \bibinfo{person}{Peter~H. Ditto}.} \bibinfo{year}{2015}\natexlab{}.
\newblock \showarticletitle{Conservatives report, but liberals display, greater
  happiness}.
\newblock \bibinfo{journal}{\emph{Science}} \bibinfo{volume}{347},
  \bibinfo{number}{6227} (\bibinfo{year}{2015}), \bibinfo{pages}{1243--1246}.
\newblock
\showISSN{0036-8075}
\urldef\tempurl%
\url{https://doi.org/10.1126/science.1260817}
\showDOI{\tempurl}


\bibitem[\protect\citeauthoryear{Wood and Porter}{Wood and Porter}{2019}]%
        {Wood}
\bibfield{author}{\bibinfo{person}{Thomas Wood} {and} \bibinfo{person}{Ethan
  Porter}.} \bibinfo{year}{2019}\natexlab{}.
\newblock \showarticletitle{{The elusive backfire effect: Mass attitudes’
  steadfast factual adherence}}.
\newblock \bibinfo{journal}{\emph{Political Behavior}} \bibinfo{volume}{41},
  \bibinfo{number}{1} (\bibinfo{year}{2019}), \bibinfo{pages}{135--163}.
\newblock


\bibitem[\protect\citeauthoryear{Zannettou}{Zannettou}{2021}]%
        {Zannettou}
\bibfield{author}{\bibinfo{person}{Savvas Zannettou}.}
  \bibinfo{year}{2021}\natexlab{}.
\newblock \showarticletitle{{``I Won the Election!'':An Empirical Analysis of
  Soft Moderation Interventions on Twitter}}.
\newblock \bibinfo{journal}{\emph{arXiv}}  \bibinfo{volume}{2101.07183v1}
  (\bibinfo{date}{18 January} \bibinfo{year}{2021}).
\newblock
\newblock
\shownote{\url{https://arxiv.org/pdf/2101.07183.pdf}.}


\bibitem[\protect\citeauthoryear{Zannettou, Caulfield, De~Cristofaro,
  Kourtelris, Leontiadis, Sirivianos, Stringhini, and Blackburn}{Zannettou
  et~al\mbox{.}}{2017}]%
        {Caulfield}
\bibfield{author}{\bibinfo{person}{Savvas Zannettou}, \bibinfo{person}{Tristan
  Caulfield}, \bibinfo{person}{Emiliano De~Cristofaro},
  \bibinfo{person}{Nicolas Kourtelris}, \bibinfo{person}{Ilias Leontiadis},
  \bibinfo{person}{Michael Sirivianos}, \bibinfo{person}{Gianluca Stringhini},
  {and} \bibinfo{person}{Jeremy Blackburn}.} \bibinfo{year}{2017}\natexlab{}.
\newblock \showarticletitle{The Web Centipede: Understanding How Web
  Communities Influence Each Other through the Lens of Mainstream and
  Alternative News Sources}. In \bibinfo{booktitle}{\emph{Proceedings of the
  2017 Internet Measurement Conference}} \emph{(\bibinfo{series}{IMC '17})}.
  \bibinfo{publisher}{Association for Computing Machinery},
  \bibinfo{address}{New York, NY, USA}, \bibinfo{pages}{405–417}.
\newblock
\showISBNx{9781450351188}
\urldef\tempurl%
\url{https://doi.org/10.1145/3131365.3131390}
\showDOI{\tempurl}


\bibitem[\protect\citeauthoryear{Zannettou, Sirivianos, Blackburn, and
  Kourtellis}{Zannettou et~al\mbox{.}}{2019}]%
        {Sirivianos}
\bibfield{author}{\bibinfo{person}{Savvas Zannettou}, \bibinfo{person}{Michael
  Sirivianos}, \bibinfo{person}{Jeremy Blackburn}, {and}
  \bibinfo{person}{Nicolas Kourtellis}.} \bibinfo{year}{2019}\natexlab{}.
\newblock \showarticletitle{The Web of False Information: Rumors, Fake News,
  Hoaxes, Clickbait, and Various Other Shenanigans}.
\newblock \bibinfo{journal}{\emph{J. Data and Information Quality}}
  \bibinfo{volume}{11}, \bibinfo{number}{3}, Article \bibinfo{articleno}{10}
  (\bibinfo{date}{May} \bibinfo{year}{2019}), \bibinfo{numpages}{37}~pages.
\newblock
\showISSN{1936-1955}
\urldef\tempurl%
\url{https://doi.org/10.1145/3309699}
\showDOI{\tempurl}


\bibitem[\protect\citeauthoryear{Zappavigna}{Zappavigna}{2018}]%
        {Zappavigna}
\bibfield{author}{\bibinfo{person}{M. Zappavigna}.}
  \bibinfo{year}{2018}\natexlab{}.
\newblock \bibinfo{booktitle}{\emph{Searchable Talk: Hashtags and Social Media
  Metadiscourse}}.
\newblock \bibinfo{publisher}{Bloomsbury Publishing}.
\newblock
\showISBNx{9781474292351}
\showLCCN{2018020548}
\urldef\tempurl%
\url{https://books.google.com/books?id=kDNWDwAAQBAJ}
\showURL{%
\tempurl}


\bibitem[\protect\citeauthoryear{Zellers, Holtzman, Rashkin, Bisk, Farhadi,
  Roesner, and Choi}{Zellers et~al\mbox{.}}{2020}]%
        {Zellers}
\bibfield{author}{\bibinfo{person}{Rowan Zellers}, \bibinfo{person}{Ari
  Holtzman}, \bibinfo{person}{Hannah Rashkin}, \bibinfo{person}{Yonatan Bisk},
  \bibinfo{person}{Ali Farhadi}, \bibinfo{person}{Franziska Roesner}, {and}
  \bibinfo{person}{Yejin Choi}.} \bibinfo{year}{2020}\natexlab{}.
\newblock \bibinfo{title}{Defending Against Neural Fake News}.
\newblock
\newblock
\showeprint[arxiv]{cs.CL/1905.12616}


\end{thebibliography}

\end{document}